\documentclass[11pt,fleqn]{article}
\usepackage{etex}
\usepackage{pstricks-add}
\reserveinserts{28}
\usepackage{pgf,tikz}
\usetikzlibrary{arrows}
\usepackage[framed,numbered,autolinebreaks,useliterate]{mcode}
\usepackage{algorithm}
\usepackage{algpseudocode}

\usepackage[dvips]{hyperref}  
\hypersetup{
  colorlinks   = true, 
  urlcolor     = blue, 
  linkcolor    = blue, 
  citecolor   = red 
}
\usepackage[pageref]{backref}
\usepackage{verbatim} 
\usepackage{hypernat}  
\usepackage{mathrsfs}
\usepackage{amstext}
\usepackage{amsmath,varwidth,array,ragged2e,tabularx,rotating}
\usepackage{amsfonts,amsthm,amssymb,amsbsy,bm,times}
\usepackage{dsfont} 
\usepackage{subeqnarray}
\usepackage{empheq} 
\usepackage{color} 
\usepackage{exscale}  
\usepackage{stmaryrd}  
\usepackage{epsfig}
\usepackage{subfigure}
\usepackage{cite}
\usepackage{calc}
\usepackage{CJK} 
\usepackage{multicol}
\usepackage{fancyhdr}
\usepackage[small]{caption}
\usepackage{empheq} 
\usepackage{color} 
\usepackage{picture,xcolor}
\usepackage{fancyvrb} 
\usepackage{enumerate}
\usepackage{cases}
\usepackage{varwidth}
\usepackage{subfigure}
\usepackage[dvips]{hyperref}  
\usepackage{exscale}  
\usepackage{datetime}
\usepackage{booktabs}
\usepackage{threeparttable}
\usepackage{stmaryrd}  
\newfont{\boldit}{cmbxti10} 
\allowdisplaybreaks[3]
\newtheorem{theorem}{Theorem}[section]
\newtheorem{lemma}[theorem]{Lemma}
\theoremstyle{definition}
\newtheorem{definition}[theorem]{Definition}

\newtheorem{proposition}{Proposition}[section]
\theoremstyle{remark}

\usepackage[all]{xy}
\usepackage{xhfill}
\newcommand{\xfilll}[2][1ex]{%
  \dimen0=#2\advance\dimen0 by #1%
  \leaders\hrule height \dimen0 depth -#1\hfill%
}

\newcommand{\tc}[1]{\multicolumn{1}{c}{#1}} 
\newcolumntype{L}{>{\varwidth[c]{\linewidth}}l<{\endvarwidth}}
\newcolumntype{M}{>{$}l<{$}}

\makeatletter
\def\hlinewd#1{%
\noalign{\ifnum0=`}\fi\hrule \@height #1 %
\futurelet\reserved@a\@xhline}

\makeatother

\numberwithin{equation}{section}
\renewcommand*{\backref}[1]{} 
\renewcommand*{\backrefalt}[4]{%
    \ifcase #1 (Not cited.)%
    \or        (Cited on page~#2.)%
    \else      (Cited on pages~#2.)%
    \fi}

\usepackage{tikz}
\newcommand*\circled[1]{\tikz[baseline=(char.base)]{
    \node[shape=circle,draw,inner sep=2pt] (char) {#1};}}

\makeatletter
\def\@cite#1#2{\textsuperscript{[{#1\if@tempswa , #2\fi}]}}
\makeatother
\begin{document}

\title{\huge A New Way to Factorize Linear Cameras}
\author{ Feng LU, Ziqiang CHEN\thanks{correspondence author}\\
}
\date{}
\maketitle

\begin{center}
{\bf Abstract}\\
\vspace{0.5cm}
\parbox{6.0in}
{\setlength
{\baselineskip}{0.5cm} The implementation details of factorizing the $3\times 4$ projection matrices of linear cameras into their {\em left matrix factors} and the $4\times 4$  homogeneous {\em central}(also {\em parallel} for infinite center cases) projection factors are presented in this work. Any full row rank $3\times 4$ real matrix can be factorized into such basic matrices which will be called LC factors.\\

A further extension to multiple view midpoint triangulation, for both pinhole and affine camera cases, is also presented based on such camera factorizations.
} \end{center}
\vspace{0.5 cm}
{\bf Keywords}: $KRt$ decomposition; pinhole camera; affine camera; midpoint triangulation.
\section{Introduction}

The homogeneous representation of geometric transformations has two distinguished features: elementary transformations such as {\em translation}, {\em central projection} can all be represented into square matrices, and the homogeneous coordinates of the {\em invariant} (or {\em fixed}) points of a transformation are exactly the eigenvectors of its homogeneous matrix. 

\begin{figure}[!hpt]
\begin{center}
\subfigure[Extended Desarguesian: ordinary central projection center]{
\newrgbcolor{afeeee}{0.69 0.93 0.93}
\newrgbcolor{qqzzff}{0 0.6 1}
\newrgbcolor{zzttqq}{0.6 0.2 0}
\newrgbcolor{uququq}{0.25 0.25 0.25}
\newrgbcolor{bcduew}{0.74 0.83 0.9}
\newrgbcolor{xdxdff}{0.49 0.49 1}
\psset{xunit=0.6cm,yunit=0.6cm,algebraic=true,dimen=middle,dotstyle=o,dotsize=3pt 0,linewidth=0.8pt,arrowsize=3pt 2,arrowinset=0.25}
\begin{pspicture*}(3.61,-5.85)(18.09,5.96)
\pspolygon[linecolor=afeeee,fillcolor=afeeee,fillstyle=solid,opacity=0.1](11.38,2.19)(10.46,0.14)(12.16,-0.83)
\pspolygon[linecolor=qqzzff,fillcolor=qqzzff,fillstyle=solid,opacity=0.1](11.38,2.19)(13.54,1.02)(12.16,-0.83)
\pspolygon[linecolor=zzttqq,fillcolor=zzttqq,fillstyle=solid,opacity=0.1](5.02,5.56)(3.98,-0.32)(8.3,-5.5)(9.49,0.53)
\psline[linecolor=qqzzff](11.38,2.19)(10.46,0.14)
\psline[linecolor=qqzzff](10.46,0.14)(12.16,-0.83)
\psline[linecolor=afeeee](12.16,-0.83)(11.38,2.19)
\psline[linecolor=qqzzff](11.38,2.19)(13.54,1.02)
\psline[linecolor=qqzzff](13.54,1.02)(12.16,-0.83)
\psline[linecolor=qqzzff](12.16,-0.83)(11.38,2.19)
\psline[linestyle=dashed,dash=3pt 3pt,linecolor=qqzzff](10.46,0.14)(13.54,1.02)
\psline[linecolor=zzttqq](5.02,5.56)(3.98,-0.32)
\psline[linecolor=zzttqq](3.98,-0.32)(8.3,-5.5)
\psline[linecolor=zzttqq](8.3,-5.5)(9.49,0.53)
\psline[linecolor=zzttqq](9.49,0.53)(5.02,5.56)
\psline[linewidth=0.4pt,linestyle=dashed,dash=3pt 3pt,linecolor=blue](8.59,-2.02)(4.98,-0.44)
\psline[linewidth=0.4pt,linestyle=dashed,dash=3pt 3pt,linecolor=blue](4.98,-0.44)(5.74,3.48)
\psline[linewidth=0.4pt,linestyle=dashed,dash=3pt 3pt,linecolor=blue](5.74,3.48)(6.22,1.34)
\psline[linewidth=0.4pt,linestyle=dashed,dash=3pt 3pt,linecolor=blue](6.22,1.34)(8.59,-2.02)
\psline[linewidth=0.4pt,linestyle=dashed,dash=3pt 3pt,linecolor=blue](5.74,3.48)(8.59,-2.02)
\psline[linewidth=0.4pt,linestyle=dashed,dash=3pt 3pt,linecolor=blue](4.98,-0.44)(6.22,1.34)
\psline[linestyle=dotted](4.98,-0.44)(9.39,0.03)
\psline(7.15,3.16)(17.21,0.85)
\psline[linestyle=dotted](5.74,3.48)(7.15,3.16)
\psline(9.02,-1.88)(17.21,0.85)
\psline[linestyle=dotted](8.59,-2.02)(9.02,-1.88)
\psline[linestyle=dotted](6.22,1.34)(8.88,1.22)
\rput[tl](7.54,-3.49){$\pi$ }
\psline(13.54,1.02)(17.21,0.85)
\psline(13.1,0.42)(17.21,0.85)
\psline(9.39,0.03)(10.46,0.14)
\psline[linestyle=dotted,linecolor=blue](10.46,0.14)(13.1,0.42)
\psline(8.88,1.22)(11.47,1.11)
\psline[linestyle=dotted,linecolor=xdxdff](11.47,1.11)(13.54,1.02)
\psdots[dotstyle=*,linecolor=blue](11.38,2.19)
\rput[bl](11.5,2.38){{$X_1$}}
\psdots[dotstyle=*,linecolor=blue](10.46,0.14)
\rput[bl](10,0.54){{$X_2$}}
\psdots[dotstyle=*,linecolor=blue](12.16,-0.83)
\rput[bl](12.41,-1.52){{$X_3$}}
\psdots[dotstyle=*,linecolor=blue](13.54,1.02)
\rput[bl](13.6,1.1){{$X_4$}}
\psdots[dotstyle=*](17.21,0.85)
\rput[bl](17.34,1.04){$S$}
\psdots[dotstyle=*,linecolor=uququq](5.74,3.48)
\rput[bl](4.91,3.57){{$Y_1$}}
\psdots[dotstyle=*,linecolor=uququq](6.22,1.34)
\rput[bl](6.07,0.38){{$Y_4$}}
\psdots[dotstyle=*,linecolor=uququq](8.59,-2.02)
\rput[bl](8.23,-2.71){{$Y_3$}}
\psdots[dotstyle=*,linecolor=uququq](4.98,-0.44)
\rput[bl](4.29,-0.46){{$Y_2$}}
\psdots[dotsize=1pt 0,dotstyle=*,linecolor=bcduew](11.47,1.11)
\end{pspicture*}
\label{fig:Desarguesian configuration projection}}
\subfigure[Extended Desarguesian: infinite projection center(parallel)]{
\newrgbcolor{zzttqq}{0.6 0.2 0}
\newrgbcolor{xdxdff}{0.49 0.49 1}
\newrgbcolor{qqzzff}{0 0.6 1}
\newrgbcolor{sqsqsq}{0.13 0.13 0.13}
\newrgbcolor{aqaqaq}{0.63 0.63 0.63}
\psset{xunit=0.6cm,yunit=0.6cm,algebraic=true,dimen=middle,dotstyle=o,dotsize=3pt 0,linewidth=0.8pt,arrowsize=3pt 2,arrowinset=0.25}
\begin{pspicture*}(-3.07,-3.2)(12.6,7.06)
\pspolygon[linecolor=zzttqq,fillcolor=zzttqq,fillstyle=solid,opacity=0.1](7.5,3.65)(5.57,2.1)(6.42,-1.2)(10.07,0.05)
\pspolygon[linecolor=qqzzff,fillcolor=qqzzff,fillstyle=solid,opacity=0.09](2.72,6.62)(-0.84,3.83)(1.39,-3.01)(4.48,0.61)
\pspolygon[linestyle=dashed,dash=2pt 2pt,linecolor=gray](1.42,4.76)(2.27,1.52)(1.55,-0.27)(3.42,2.42)
\pspolygon[linestyle=dashed,dash=2pt 2pt,linecolor=gray](1.55,-0.27)(1.42,4.76)(2.27,1.52)
\pspolygon[linecolor=zzttqq,fillcolor=zzttqq,fillstyle=solid,opacity=0.13](6.42,-1.2)(5.57,2.1)(7.5,3.65)
\psline[linecolor=zzttqq](7.5,3.65)(5.57,2.1)
\psline[linecolor=zzttqq](5.57,2.1)(6.42,-1.2)
\psline[linecolor=zzttqq](6.42,-1.2)(10.07,0.05)
\psline[linecolor=zzttqq](10.07,0.05)(7.5,3.65)
\psline[linecolor=xdxdff](6.42,-1.2)(7.5,3.65)
\psline[linecolor=qqzzff](2.72,6.62)(-0.84,3.83)
\psline[linecolor=qqzzff](-0.84,3.83)(1.39,-3.01)
\psline[linecolor=qqzzff](1.39,-3.01)(4.48,0.61)
\psline[linecolor=qqzzff](4.48,0.61)(2.72,6.62)
\psline[linestyle=dashed,dash=3pt 3pt,linecolor=gray](2.27,1.52)(1.55,-0.27)
\psline[linestyle=dashed,dash=2pt 2pt,linecolor=gray](1.55,-0.27)(3.42,2.42)
\psline[linestyle=dashed,dash=3pt 3pt,linecolor=gray](3.42,2.42)(1.42,4.76)
\psline[linecolor=xdxdff](1.42,4.76)(13.21,2.6)
\psline[linecolor=xdxdff](-3.51,5.66)(0.55,4.92)
\psline[linecolor=xdxdff](-3.48,3.76)(-0.65,3.24)
\psline[linecolor=xdxdff](-3.52,2.54)(-0.23,1.94)
\psline[linecolor=xdxdff](-3.33,0.59)(0.44,-0.1)
\psline[linecolor=xdxdff](1.55,-0.27)(13.52,-2.5)
\psline[linecolor=xdxdff](10.07,0.05)(13.33,-0.54)
\psline[linecolor=xdxdff](2.27,1.52)(6.32,0.74)
\psline[linecolor=xdxdff](3.42,2.42)(5.57,2.1)
\psline[linecolor=xdxdff](9.06,1.46)(13.36,0.68)
\psline[linestyle=dashed,dash=2pt 2pt,linecolor=gray](1.55,-0.27)(1.42,4.76)
\psline[linestyle=dashed,dash=3pt 3pt,linecolor=gray](1.42,4.76)(2.27,1.52)
\psline[linecolor=zzttqq](6.42,-1.2)(5.57,2.1)
\psline[linecolor=zzttqq](5.57,2.1)(7.5,3.65)
\psline[linecolor=zzttqq](7.5,3.65)(6.42,-1.2)
\rput[tl](10.24,2.77){$S_{\infty }$}
\rput[tl](-1.79,5.04){$S_{\infty }$}
\psline[linewidth=0.4pt]{->}(11.19,2)(12.39,1.78)
\psline[linewidth=0.4pt]{->}(-2.01,4.41)(-3,4.59)
\rput[tl](1.4,-1.54){$\pi$}
\psline[linestyle=dashed,dash=3pt 3pt,linecolor=aqaqaq](2.27,1.52)(3.42,2.42)
\psdots[dotstyle=*](7.5,3.65)
\rput[bl](7.64,3.84){$X_1$}
\psdots[dotstyle=*](5.57,2.1)
\rput[bl](5.25,2.56){$X_2$}
\psdots[dotstyle=*](6.42,-1.2)
\rput[bl](6.17,-2){$X_3$}
\psdots[dotstyle=*](10.07,0.05)
\rput[bl](10.21,0.32){$X_4$}
\psdots[dotstyle=*,linecolor=sqsqsq](1.42,4.76)
\rput[bl](1.55,4.95){\sqsqsq{$Y_1$}}
\psdots[dotstyle=*,linecolor=sqsqsq](2.27,1.52)
\rput[bl](2.28,2.44){\sqsqsq{$Y_4$}}
\psdots[dotstyle=*,linecolor=sqsqsq](1.55,-0.27)
\rput[bl](1.52,-0.99){\sqsqsq{$Y_3$}}
\psdots[dotstyle=*,linecolor=sqsqsq](3.42,2.42)
\rput[bl](3.54,2.59){\sqsqsq{$Y_2$}}
\psdots[dotsize=1pt 0,dotstyle=*,linecolor=lightgray](6.32,0.74)
\end{pspicture*}
\label{fig:Desarguesian configuration parallel projection}}
\end{center}
\caption{Extended Desarguesian configuration for central ({\em parallel} for infinite center) projection}
\end{figure}
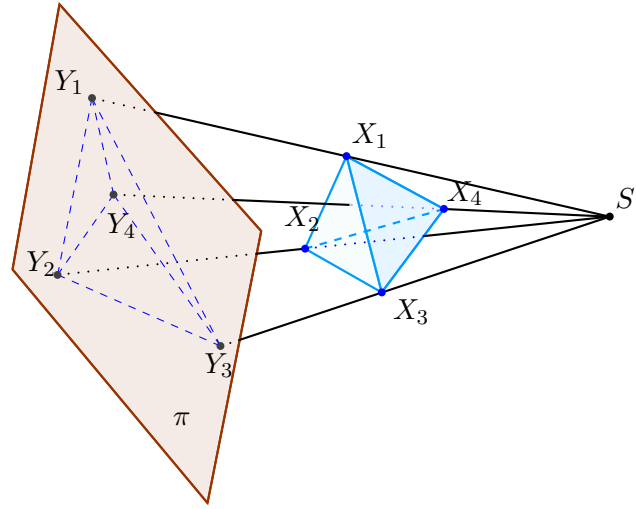
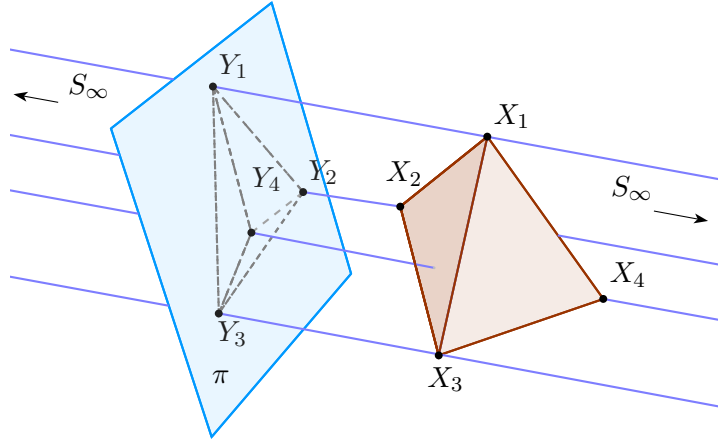

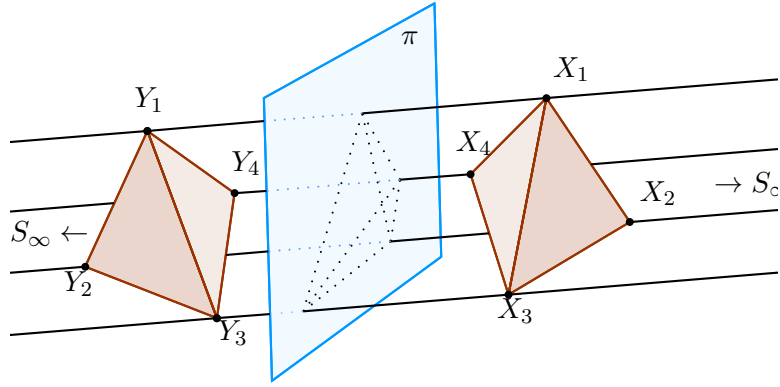
\begin{figure}[!hpbt]
\begin{center}
\newrgbcolor{zzttqq}{0.6 0.2 0}
\newrgbcolor{qqzzff}{0 0.6 1}
\newrgbcolor{wwzzff}{0.4 0.6 1}
\newrgbcolor{zzccff}{0.6 0.8 1}
\psset{xunit=0.6cm,yunit=0.6cm,algebraic=true,dotstyle=o,dotsize=3pt 0,linewidth=0.8pt,arrowsize=3pt 2,arrowinset=0.25}
\begin{pspicture*}(-3,-4)(14,6)
\pspolygon[linecolor=zzttqq,fillcolor=zzttqq,fillstyle=solid,opacity=0.2](8.88,3.19)(10.73,0.44)(8.04,-1.17)
\pspolygon[linecolor=zzttqq,fillcolor=zzttqq,fillstyle=solid,opacity=0.1](8.88,3.19)(7.2,1.5)(8.04,-1.17)
\pspolygon[linecolor=zzttqq,fillcolor=zzttqq,fillstyle=solid,opacity=0.2](0.04,2.46)(-1.34,-0.55)(1.58,-1.69)
\pspolygon[linecolor=zzttqq,fillcolor=zzttqq,fillstyle=solid,opacity=0.1](0.04,2.46)(1.97,1.08)(1.58,-1.69)
\pspolygon[linecolor=qqzzff,fillcolor=qqzzff,fillstyle=solid,opacity=0.05](2.62,3.19)(6.4,5.29)(6.55,-0.33)(2.8,-3.08)
\psline[linecolor=zzttqq](8.88,3.19)(10.73,0.44)
\psline[linecolor=zzttqq](10.73,0.44)(8.04,-1.17)
\psline[linecolor=zzttqq](8.04,-1.17)(8.88,3.19)
\psline[linecolor=zzttqq](8.88,3.19)(7.2,1.5)
\psline[linecolor=zzttqq](7.2,1.5)(8.04,-1.17)
\psline[linecolor=zzttqq](8.04,-1.17)(8.88,3.19)
\psline[linecolor=zzttqq](0.04,2.46)(-1.34,-0.55)
\psline[linecolor=zzttqq](-1.34,-0.55)(1.58,-1.69)
\psline[linecolor=zzttqq](1.58,-1.69)(0.04,2.46)
\psline[linecolor=zzttqq](0.04,2.46)(1.97,1.08)
\psline[linecolor=zzttqq](1.97,1.08)(1.58,-1.69)
\psline[linecolor=zzttqq](1.58,-1.69)(0.04,2.46)
\psline[linecolor=qqzzff](2.62,3.19)(6.4,5.29)
\psline[linecolor=qqzzff](6.4,5.29)(6.55,-0.33)
\psline[linecolor=qqzzff](6.55,-0.33)(2.8,-3.08)
\psline[linecolor=qqzzff](2.8,-3.08)(2.62,3.19)
\psline[linestyle=dotted](4.8,2.85)(5.64,1.38)
\psline[linestyle=dotted](5.64,1.38)(5.43,0.01)
\psline[linestyle=dotted](3.49,-1.54)(5.43,0.01)
\psline[linestyle=dotted](4.8,2.85)(3.49,-1.54)
\psline[linestyle=dotted](4.8,2.85)(5.43,0.01)
\psline[linestyle=dotted](5.64,1.38)(3.49,-1.54)
\psline(5.64,1.38)(7.2,1.5)
\psline(4.8,2.85)(8.88,3.19)
\psline(3.49,-1.54)(8.04,-1.17)
\psline(5.43,0.01)(7.61,0.19)
\psline(1.58,-1.69)(2.76,-1.6)
\psline(1.78,-0.29)(2.72,-0.21)
\psline(1.97,1.08)(2.68,1.13)
\psline(0.04,2.46)(2.63,2.68)
\psline(-6.01,-2.31)(1.58,-1.69)
\psline(-6.16,-0.94)(-1.34,-0.55)
\psline(-6.28,1.95)(0.04,2.46)
\psline(-6.36,0.4)(-0.7,0.86)
\psline(8.88,3.19)(18.18,3.94)
\psline(8.04,-1.17)(18.67,-0.3)
\psline(10.73,0.44)(18.6,1.08)
\psline(9.87,1.72)(18.42,2.42)
\psline[linestyle=dotted,linecolor=wwzzff](2.63,2.68)(4.8,2.85)
\psline[linestyle=dotted,linecolor=wwzzff](2.68,1.13)(5.64,1.38)
\psline[linestyle=dotted,linecolor=wwzzff](2.72,-0.21)(5.43,0.01)
\psline[linestyle=dotted,linecolor=zzccff](2.76,-1.6)(3.49,-1.54)
\rput[tl](5.66,4.67){$\pi$}
\rput[tl](12.6,1.55){$\to S_{\infty}$}
\rput[tl](-3,0.45){$ S_{\infty} \leftarrow$}
\psdots[dotstyle=*](8.88,3.19)
\rput[bl](9.03,3.56){$X_1$}
\psdots[dotstyle=*](10.73,0.44)
\rput[bl](10.95,0.81){$X_2$}
\psdots[dotstyle=*](8.04,-1.17)
\rput[bl](7.79,-1.79){$X_3$}
\psdots[dotstyle=*](7.2,1.5)
\rput[bl](6.87,1.92){$X_4$}
\psdots[dotstyle=*](0.04,2.46)
\rput[bl](-0.25,2.89){$Y_1$}
\psdots[dotstyle=*](-1.34,-0.55)
\rput[bl](-1.81,-1.19){$Y_2$}
\psdots[dotstyle=*](1.58,-1.69)
\rput[bl](1.61,-2.28){$Y_3$}
\psdots[dotstyle=*](0.04,2.46)
\psdots[dotstyle=*](1.97,1.08)
\rput[bl](1.9,1.48){$Y_4$}
\psdots[dotstyle=*](1.58,-1.69)
\end{pspicture*}
\end{center}
\caption{Extended Desarguesian configuration for reflection}
\label{fig:Desarguesian configuration reflection}
\end{figure}

Taking advantage of these features, Chen~\cite{Chen2000, Chen2005} proposed the concept {\em stereohomology} which unified a series of elementary geometric operators into modified Householder elementary matrices\cite[pp.1-3]{Householder-8}~(Table~\ref{classification table}~\cite{LuChen2013}). The definition of stereohomology is based on an extension to Desargues theorem. Figures~\ref{fig:Desarguesian configuration projection}, \ref{fig:Desarguesian configuration parallel projection},and~\ref{fig:Desarguesian configuration reflection} and illustrate the {\em extended Desarguesian configurations} for typical {\em central projection}, {\em parallel projection},and {\em reflection}, respectively.

A major advantage of stereohomology representation should be its {\em explicit parameterization}. For example, a 3D orthographic reflection has at most three and a 3D central(also {\em parallel} for infinite center cases) projection has at most six independent parameters while their homogeneous matrices have 15 entries: $4\times 4$ matrices up to a scale factor. In the modified Householder elementary matrix forms~(Table~\ref{classification table}~\cite{LuChen2013}), these homogeneous matrices can be represented in a form with exactly the same independent explicit parameter number as requisite, i.e., not more than 3 for reflections and 6 for central(5 for parallel) projections.

Since every linear camera's inherent geometric operation is {\em central}(also {\em parallel} for infinite centers) projection, it becomes interesting how to extract the $4\times 4$ projection matrix from a 3$\times$4 linear camera. LC factorization is such a procedure which factorizes a linear camera into its {\em left camera factors} and a $4\times 4$ {\em central}(also {\em parallel} for infinite centers) projection matrix.

In this work, we present the implementation details of LC factorization for both pinhole and affine cameras,  and apply the so-called LC factorization to generalize the multiple view mid-point triangulation(also called {\em symmedian point} triangulation~\cite{iterativeSolver2014}) for both pinhole and affine cameras.
\pagebreak

This paper is organized as follows: in section~\ref{LC factorization section} we introduce the implementation details and the pseudo-codes of the LC factorizations for both pinhole and affine cameras;  section~\ref{symmedian point triangulation section} presents symmdian point triangulation which is a multiple-view generalization of mid-point triangulation based on LC factorization; we put in the first part of Appendix section $RQ$ decomposition related discussion.

\section{LC Factorizations}\label{LC factorization section}

We need to use some elementary operators as factors. The {\em translation}, {\em shearing}, {\em scaling} and {\em rotation} operators with explicit parameters are as simple as the conventional ones. The {\em reflection}, {\em central projection} and {\em parallel projection} operators are already available with both algebraic definitions and formulations with explicit parameters~\cite{Chen2000, Chen2005, LuChen2013}(see Table~\ref{classification table}).

By saying linear camera in this paper, we mean any $3\times 4$ real matrix with full row rank. Since a linear camera converts 3D scenes into 2D images, the homogeneous matrix of a camera is in essence 3$\times$4 in dimensions and is different from the square versions of central(parallel) projections in algebraic projective geometry. We shall introduce another two {\em basic} matrices appeared in~\cite{Chen2005}: {\em cutting} and {\em augmenting} so as to keep the major matrix factors square in shape and to keep their physical meaning clear in algebraic projective geometry.

\begin{definition}[Elementary cutting matrix]\label{elementary cutting}
An elementary cutting matrix is an identity square matrix with any row or column cut off.
\end{definition}

\begin{definition}[Elementary augmenting matrix]\label{elementary augmenting}
Insert an all-zero row or column into an identity square matrix, then we obtain an elementary augmenting matrix.
\end{definition}

For example, the cutting matrix in equation~\eqref{eqn:3D to 2D}(left), obtained by removing the third row of a $4\times 4$ identity matrix, transforms a 3D point $(x,y,0,1)^T$ on $xoy$ axis plane into the 2D one: $(x,y,1)^T$ where the 2D coordinate system coincides with the $xoy$ axis plane of the 3D ones, while the augmenting matrix (right), obtained by inserting an all-zero row vector right after the second row of a $3\times 3$ identity matrix, transforms the 2D point into the 3D point conversely:

\begin{align}\label{eqn:3D to 2D}
\left[ {\begin{array}{*{20}c}
   x  \\
   y  \\
   1  \\
 \end{array} } \right] = \underbrace{\left[ {\begin{array}{*{20}c}
   1 & 0 & 0 & 0  \\
   0 & 1 & 0 & 0  \\
   0 & 0 & 0 & 1  \\
 \end{array} } \right]}_{\text{Cutting}} \left[ {\begin{array}{*{20}c}
   x  \\
   y  \\
   0  \\
   1  \\
 \end{array} } \right];\qquad
\left[ {\begin{array}{*{20}c}
   x  \\
   y  \\
   0  \\
   1  \\
 \end{array} } \right] = \underbrace{\left[ {\begin{array}{*{20}c}
   1 & 0 & 0  \\
   0 & 1 & 0  \\
   0 & 0 & 0  \\
   0 & 0 & 1  \\
 \end{array} } \right]}_{\text{Augmenting}} \left[ {\begin{array}{*{20}c}
   x  \\
   y  \\
   1  \\
 \end{array} } \right]
\end{align}

The cutting--augmenting matrix pair for $xoy$ axis plane in equation~\eqref{eqn:3D to 2D} are also transpose to each other. Similar rules hold for the cutting--augmenting pairs of $yoz$ and $xoz$ axis planes.

Let us first introduce the lemma below, on which we will depend to establish a critical {\em reflection} factor:

\begin{lemma}\label{lemma:bisection planes}
 Given two planes $\pi_1:$ $a_1 x + b_1 y+ c_1 z +d_1 =0$ and $\pi_2:$ $a_2 x+ b_2y +c_2z+d_2=0$ with a dihedral angle $\omega$ between them, where $a_i^2+b_i^2+c_i^2=1$ $(i=1,2)$ and $\sum\limits_{k=a,b,c}\left(k_1\cdot k_2\right)=-\cos\omega$, two planes having dihedral angle $\theta$ to $\pi_1$ have the following equations when $\pi_1\ne\pi_2$:
\begin{equation}\label{3D rotation hyperplanes}
\left\{ \begin{array}{*{20}{c}}
\pi_{\omega-\theta}:\sin\left(\omega-\theta\right)\left(a_1x + b_1y + c_1z +d_1\right)+\sin\theta \left(a_2x + b_2y+c_2z+d_2\right)=0\\
\pi_{\omega+\theta}:\sin\left(\omega+\theta\right)\left(a_1x + b_1y + c_1z +d_1\right)-\sin\theta \left(a_2x + b_2y+c_2z+d_2\right)=0
\end{array}
\right.
\end{equation}
Consequently $\pi_{\omega\pm\theta}$ become the ``bisection planes" of $\pi_1$ and $\pi_2$  with equation $\pi_1\pm \pi_2$ When $\theta=\omega/2$,.
\end{lemma}

\subsection{LC Factorization of Pinhole Cameras }\label{LC factorization pinhole}
\noindent  As is well known that the homogeneous coordinate of a camera center is exactly the one dimensional right null-space of the camera matrix. This holds for both pinhole and affine cameras. Any linear camera has their {\em KRt} decomposition~\cite{Faugeras:2001, ZissermanHartley-5,Ma:2003:AnInvitation,Faugeras:1993} by $RQ$ decomposition~\cite{ZissermanHartley-5,VGG2012}. LC factorization keeps compatibility with the conventional methods, i.e., the physical meaning of key parameters will remain almost the same and {\em KRt} can easily be deduced from LC factors.


Then we introduce the geometric meaning of the matrix factors in equation~\eqref{LC camera representation0} in details. Note that such factors and their orders as in equation~\eqref{LC camera representation0} are not unique. In this paper, we only discuss the cases of LC factorization defined by equation~\eqref{LC camera representation0} and~\eqref{eqn:LC factorization affine}.

{\begin{equation}\label{LC camera representation0}
\begin{gathered}
\boldsymbol{P}= \underbrace {\hphantom{A} \boldsymbol{ L }\hphantom{A}}_{\circled{8}\text{~\texttildelow~}\circled{2}} \cdot \underbrace{\hphantom{A}\mathscr{C}\hphantom{A}}_{\circled{1}} = \underbrace {\left[ {\begin{array}{*{20}{c}}
  1&0&{  u} \\
  0&1&{  v} \\
  0&0&1
\end{array}} \right]}_{\circled{8}}\cdot\underbrace {\left[ {\begin{array}{*{20}{c}}
  1&\tau&{  0} \\
  0&1&{  0} \\
  0&0&1
\end{array}} \right]}_{\circled{7}}\cdot \underbrace {\left[ {\begin{array}{*{20}{c}}
  \sigma&0&{  0} \\
  0&1&{  0} \\
  0&0&1
\end{array}} \right]}_{\circled{6}}\cdot\underbrace {\left[ {\begin{array}{*{20}{c}}
  {\cos \alpha }&{\sin \alpha }&0 \\
  { - \sin \alpha }&{\cos \alpha }&0 \\
  0&0&1
\end{array}} \right]}_{\circled{5}} \hfill \\ \hphantom{A}\cdot\underbrace {\left[ {\begin{array}{*{20}{c}}
  1&0&{ - u_s} \\
  0&1&{ - v_s} \\
  0&0&1
\end{array}} \right]}_{\circled{4}}
  \cdot\underbrace {\left[ {\begin{array}{*{20}{c}}
  1&0&0&0 \\
  0&1&0&0 \\
  0&0&0&1
\end{array}} \right]}_{\circled{3}}
\cdot\underbrace {\left( {I - 2\frac{{\left( {{S_R}} \right){{\left( {{\pi _R}} \right)}^ T }}}{{{{\left( {{S_R}} \right)}^ T }\left( {{\pi _R}} \right)}}} \right)}_{\circled{2}}
\underbrace {\left( {I - \frac{{\left( s \right){{\left( \pi  \right)}^ T }}}{{{{\left( s \right)}^ T }\left( \pi  \right)}}} \right)}_{\circled{1}}\hfill \\
\end{gathered}
\end{equation}}
\noindent where $\mathscr{C}$ is the inherent {\em central projection}, while $\boldsymbol L$ represents {\em left matrix factors}, and parameters are:{\small
\begin{align*}\label{LC factorization parameters}
\hphantom{a}\left(s\right) &=\left(x_s,y_s,z_s,1\right)^T ,\text{ is the homogeneous coordinate of the inherent projection center; } \hfill\\
\hphantom{a}\left(\pi\right) &= \left( r \cos \theta, r \sin \theta, \sqrt{1-r^2}, d \right)^T ,\text{ where: }d=f-\left(r x_s \cos \theta+ r y_s \sin \theta+ z_s \sqrt{1-r^2} \right), \text{ image plane;}\hfill\\
\left(\pi_R \right) &= \left( r \cos \theta, r \sin \theta, \sqrt{1-r^2}+1, d\right)^T , \text{bisection between }(\pi)\text{ and $xoy$ plane $(0,0,1,0)^T$ per Lemma~\ref{lemma:bisection planes}}\hfill\\
\left(S_R \right)& = \left( r \cos \theta, r \sin \theta, \sqrt{1-r^2}+1, 0\right)^T,  \text{ normal direction of mirror plane $\pi_R$;}  \hfill \\
\hphantom{A}u_s &= {x_s}-\dfrac{r(f+{z_s})}{1+\sqrt{1-r^2}}\cos\theta, \text{ is the }x\text{ coordinate of reflected } \left(s\right) \text{ by \circled{2}}; \hfill \\
\hphantom{A}v_s &= {y_s}-\dfrac{r(f+{z_s})}{1+\sqrt{1-r^2}}\sin\theta, \text{ is the }y \text{ coordinate of reflected } \left(s\right).\hfill
\end{align*}

In equation~\eqref{LC camera representation0}, matrix \circled{1} is the 4$\times$4 homogeneous {\em central projection} in the 3D world coordinate system per definition in~\cite{LuChen2013}, which is uniquely determined by its projection center $S$ and the projection plane $\pi$ as in Figure~\ref{fig:LC factorization geometric transforms}, where ${\boldsymbol y} = (y_1, y_2, y_3, 1)^\top$ is a 3D image point on $\pi$ of the scene point $Y$  via the central projection \circled{1}.

\begin{figure}[!hpt]
\begin{center}
\newrgbcolor{xdxdff}{0.49 0.49 1}
\newrgbcolor{zzttqq}{0.6 0.2 0}
\newrgbcolor{qqzzff}{0 0.6 1}
\newrgbcolor{wwzzqq}{0.4 0.6 0}
\newrgbcolor{zzwwqq}{0.6 0.4 0}
\newrgbcolor{qqzzcc}{0 0.6 0.8}
\newrgbcolor{zzccff}{0.6 0.8 1}
\newrgbcolor{yqqqqq}{0.5 0 0}
\psset{xunit=1.0cm,yunit=1.0cm,algebraic=true,dimen=middle,dotstyle=o,dotsize=3pt 0,linewidth=0.8pt,arrowsize=3pt 2,arrowinset=0.25}
\begin{pspicture*}(-0.4,-2.92)(8.69,5.76)
\pspolygon[linecolor=zzttqq,fillcolor=zzttqq,fillstyle=solid,opacity=0.1](4.58,1.57)(3.82,5.64)(6.44,4.14)(7.19,-0.06)
\pspolygon[linecolor=qqzzff,fillcolor=qqzzff,fillstyle=solid,opacity=0.2](4.58,1.57)(0.42,2.35)(2.57,0.38)(7.19,-0.06)
\pspolygon[linecolor=zzwwqq](0.35,-0.97)(0.72,-0.96)(0.52,-1.22)(0.16,-1.22)
\pspolygon[linecolor=zzttqq,fillcolor=zzttqq,fillstyle=solid,opacity=0.1](0.16,-1.22)(1.87,1.02)(2.57,0.38)(7.17,-0.06)(6.38,-1.19)
\pspolygon[linestyle=dotted,linecolor=zzttqq,fillcolor=zzttqq,fillstyle=solid,opacity=0.1](4.58,1.57)(8.42,1.64)(7.17,-0.06)
\psline{->}(0.16,-1.22)(8.36,-1.18)
\psline{->}(0.16,-1.22)(0.16,4.84)
\psline[linecolor=zzttqq](4.58,1.57)(3.82,5.64)
\psline[linecolor=zzttqq](3.82,5.64)(6.44,4.14)
\psline[linecolor=zzttqq](6.44,4.14)(7.19,-0.06)
\psline[linestyle=dotted,linecolor=zzttqq](7.19,-0.06)(4.58,1.57)
\psline(8.14,4.56)(6.52,3.7)
\psline(3.18,1.92)(5.47,3.14)
\psline[linecolor=qqzzff](4.58,1.57)(0.42,2.35)
\psline[linecolor=qqzzff](0.42,2.35)(2.57,0.38)
\psline[linecolor=qqzzff](2.57,0.38)(7.19,-0.06)
\psline[linecolor=qqzzff](7.19,-0.06)(4.58,1.57)
\rput[tl](7.64,-1.3){$x$}
\rput[tl](0.46,4.01){$z$}
\rput[tl](2.57,2.7){$y$}
\rput[tl](0.26,-1.4){$O$}
\parametricplot[linecolor=wwzzqq]{1.7472715663003966}{3.0465364528048062}{1*0.59*cos(t)+0*0.59*sin(t)+7.19|0*0.59*cos(t)+1*0.59*sin(t)+-0.06}
\rput[tl](6.55,0.85){$\theta$}
\rput[tl](6,-0.36){$\theta$}
\rput[tl](4.32,4.88){$\pi$}
\psline[linecolor=zzwwqq](0.35,-0.97)(0.72,-0.96)
\psline[linecolor=zzwwqq](0.72,-0.96)(0.52,-1.22)
\psline[linecolor=zzwwqq](0.52,-1.22)(0.16,-1.22)
\psline[linecolor=zzwwqq](0.16,-1.22)(0.35,-0.97)
\rput[tl](1.16,-0.56){$x$-$o$-$y$}
\rput[tl](5.08,0.59){$\pi_{\omega-\theta}$}
\rput[tl](7.66,1.53){$\pi_1$}
\psline(3.18,1.92)(3.02,1.32)
\psline[linestyle=dotted](2.89,0.84)(3.02,1.32)
\psline[linecolor=zzccff](3.02,1.32)(5.51,0.99)
\rput[tl](5.77,1.28){$l$}
\psline[linecolor=qqzzff](6.84,-2.43)(5.35,-1.19)
\psline[linestyle=dotted,linecolor=blue](4.47,-0.47)(5.35,-1.19)
\psline[linecolor=qqzzff](3.57,0.28)(4.47,-0.47)
\psline[linestyle=dotted,linecolor=blue](2.89,0.84)(3.57,0.28)
\psline[linecolor=yqqqqq](6.38,-1.19)(8.42,1.64)
\psline[linecolor=yqqqqq](6.89,1.61)(8.42,1.64)
\psline[linestyle=dotted,linecolor=zzttqq](4.58,1.57)(2.26,1.53)
\psline(0.16,-1.22)(1.87,1.02)
\psline{->}(2.58,1.95)(3.2,2.76)
\psline[linestyle=dotted,linecolor=zzttqq](1.87,1.02)(2.58,1.95)
\psline[linestyle=dotted,linecolor=zzttqq](0.16,-1.22)(1.87,1.02)
\psline[linestyle=dotted,linecolor=zzttqq](7.17,-0.06)(6.38,-1.19)
\psline[linestyle=dotted,linecolor=zzttqq](6.38,-1.19)(0.16,-1.22)
\psline[linestyle=dotted,linecolor=zzttqq](8.42,1.64)(7.17,-0.06)
\psline[linecolor=qqzzff](0.42,2.35)(2.57,0.38)
\psline[linecolor=qqzzff](2.57,0.38)(7.17,-0.06)
\psline[linewidth=0.4pt,linestyle=dotted](6.5,-0.61)(7.19,-0.06)
\parametricplot[linecolor=blue]{3.0465364528048062}{3.808685342520821}{1*0.88*cos(t)+0*0.88*sin(t)+7.19|0*0.88*cos(t)+1*0.88*sin(t)+-0.06}
\parametricplot[linecolor=blue]{3.0465364528048062}{3.808685342520821}{1*0.66*cos(t)+0*0.66*sin(t)+7.19|0*0.66*cos(t)+1*0.66*sin(t)+-0.06}
\psdots[dotstyle=*](0.16,-1.22)
\psdots[dotstyle=*,linecolor=xdxdff](7.19,-0.06)
\psdots[dotstyle=*](3.18,1.92)
\rput[bl](3.39,1.68){$S$}
\psdots[dotstyle=*](8.14,4.56)
\rput[bl](8.23,4.67){$Y$}
\psdots[dotstyle=*](5.47,3.14)
\rput[bl](5.55,3.27){$\boldsymbol y$}
\psdots[dotsize=1pt 0,dotstyle=*,linecolor=qqzzcc](3.02,1.32)
\psdots[dotstyle=*](2.89,0.84)
\rput[bl](2.39,0.85){$S^*$}
\psdots[dotstyle=*](6.84,-2.43)
\rput[bl](6.22,-2.76){$Y^*$}
\psdots[dotstyle=*](4.47,-0.47)
\rput[bl](4.21,-1){${\boldsymbol y}^*$}
\end{pspicture*}
\end{center}
\caption{How to reflect an image plane $\pi$ of a central projection into $xoy$ axis plane}
\label{fig:LC factorization geometric transforms}
\end{figure}
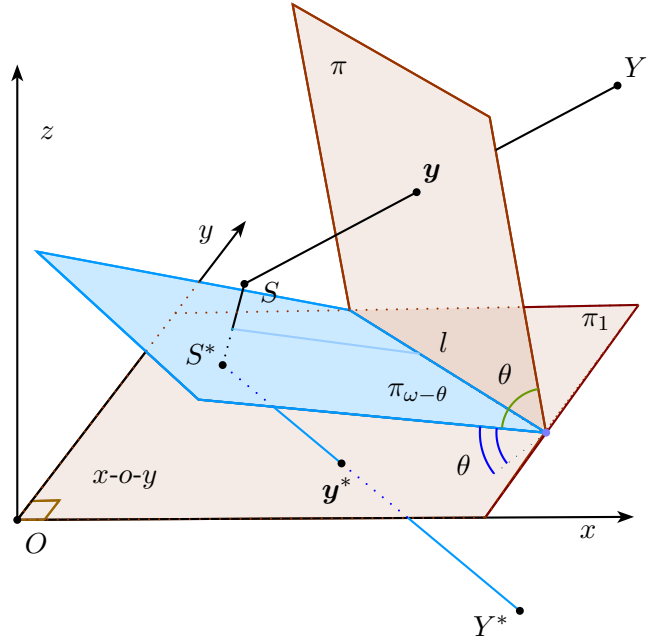

In order to obtain the $3\times4$ dimensional {\em three2two} matrix which converts 3D images into 2D images, algebraically we concatenate matrices~\circled{8}~\texttildelow~\circled{2} by matrix multiplication; conversely, by concatenating the inverses of square matrices in~\circled{2}~\texttildelow~\circled{8} and replace \circled{3} with {\em elementary augmenting}~\cite{LuChen2013} in reversed sequence, we obtain the $4\times 3$ {\em two2three} matrix which converts 2D images into 3D ones.

Since elementary cutting \circled{3} only works for 3D points on 3D axis planes~\eqref{eqn:3D to 2D}, we first transform $\pi$ into a desired axis plane, e.g., $xoy$ ($0,0,1,0$)$^T$, by a distance-preserving transformation \circled{2} before the \circled{3}(cutting) operation. Let the 3D image plane ($\pi$)=($a,b,c,d$)$^T$ as in figure~\ref{fig:LC factorization geometric transforms}.

We choose a reflection~\circled{2} with the bisection plane determined per Lemma~\ref{lemma:bisection planes} as the mirror plane, then such a conversion is as illustrated in Figure~\ref{fig:LC factorization geometric transforms}. Denote $xoy$ axis plane as $\pi_1$=($0,0,1,0$)$^T$, and ($\pi$) = $\left(r\cos\theta,r\sin\theta,\sqrt{1-r^2},d\right)^T$ ($0\leqslant r \leqslant 1$), then either of the two bisection planes can be used as the mirror plane of reflection. We can chose $\pi_{\omega-{\theta}}$ as ($\pi_R$) of~\circled{2} in equation~\eqref{LC camera representation0}, of which the homogeneous coordiante is $\left(r\cos\theta,r\sin\theta,\sqrt{1-r^2}+1,d\right)^T$, and ($S_R$) is the normal direction of ($\pi_R$)~\cite{LuChen2013}. In Figure~\ref{fig:LC factorization geometric transforms}, the 3D image point ${\boldsymbol y}$ is thus reflected into a 3D image ${\boldsymbol y}^*$ on the axis plane $xoy$. Note that the projection center $S$ is also reflected into $S^*$ at the same time.

Then by elementary cutting(\circled{3} in~\eqref{LC camera representation0}), the 3D image point ${\boldsymbol y}^*$ is converted into a 2D one~\eqref{eqn:3D to 2D}.

Translation \circled{4} in equation~\eqref{LC camera representation0} is to assure the orthographic projection of the reflected $S^*$ by \circled{2} is exactly at the origin of the 2D image reference system, where $u_s$ and $v_s$ are the quotients of the first and second coordinate components divided by the fourth component of ($s$) in \circled{1} after being reflected by \circled{2}. This additional \circled{4}, keeps the physical meanings of the intrinsic parameters by LC factorization compatible with conventional camera decomposition approaches.

\begin{figure}[!hpt]
\begin{center}
\newrgbcolor{zzttqq}{0.6 0.2 0}
\newrgbcolor{xdxdff}{0.49 0.49 1}
\psset{xunit=.75cm,yunit=.75cm,algebraic=true,dotstyle=o,dotsize=3pt 0,linewidth=0.8pt,arrowsize=3pt 2,arrowinset=0.25}
\begin{pspicture*}(1.5,-2)(10.5,5)
\pspolygon[linecolor=zzttqq,fillcolor=zzttqq,fillstyle=solid,opacity=0.05](7.36,4.26)(9.48,2.44)(9.4,-1.5)(7.28,0.28)
\pspolygon[linecolor=zzttqq,fillcolor=zzttqq,fillstyle=solid,opacity=0.05](2.26,4.24)(4.38,2.42)(4.3,-1.52)(2.18,0.26)
\psline[linecolor=zzttqq](7.36,4.26)(9.48,2.44)
\psline[linecolor=zzttqq](9.48,2.44)(9.4,-1.5)
\psline[linecolor=zzttqq](9.4,-1.5)(7.28,0.28)
\psline[linecolor=zzttqq](7.28,0.28)(7.36,4.26)
\rput[tl](7.58,3.56){$\pi_1$}
\psline[linecolor=zzttqq](2.26,4.24)(4.38,2.42)
\psline[linecolor=zzttqq](4.38,2.42)(4.3,-1.52)
\psline[linecolor=zzttqq](4.3,-1.52)(2.18,0.26)
\psline[linecolor=zzttqq](2.18,0.26)(2.26,4.24)
\rput[tl](2.56,3.24){$\pi_2$}
\psline(8.53,3.26)(9.82,4.14)
\psline(7.95,2.87)(4.34,0.41)
\psline(1.98,-1.2)(3.54,-0.13)
\psline[linestyle=dotted](4.34,0.41)(3.54,-0.13)
\psline[linestyle=dotted](8.53,3.26)(7.95,2.87)
\psdots[dotstyle=*](9.82,4.14)
\rput[bl](9.42,4.26){$X$}
\psdots[dotstyle=*](5.83,1.42)
\rput[bl](5.56,1.64){$S$}
\psdots[dotstyle=*,linecolor=xdxdff](7.95,2.87)
\rput[bl](8.04,2.48){{$x_1$}}
\psdots[dotstyle=*,linecolor=xdxdff](3.54,-0.13)
\rput[bl](3.64,-0.5){{$x_2$}}
\end{pspicture*}
\end{center}
\caption{Two possibilities of the central projection \protect\circled{1} in~\eqref{LC factorization pinhole}}
\label{fig:two possibilities LC factorization}
\end{figure}
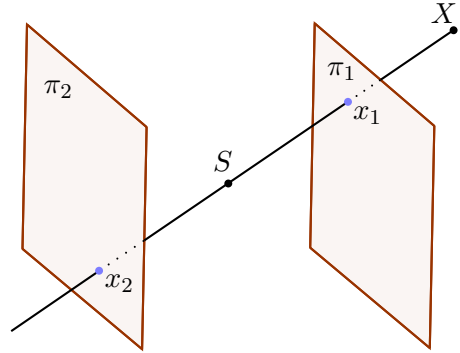

Further transformations in sequence, i.e., rotation \circled{5} around the 2D image plane origin, scaling \circled{6} and shearing \circled{7} representing aspect ratio and skewness respectively, and the further translation \circled{8}, are employed so that the customized 2D image coordinate system can be arbitrary.

There are 11 independent parameters in total in equation~\eqref{LC camera representation0}, which are: oriented focal length (1) $f$, which can be positive or negative and means the two equivalent possibilities of central projection(see Figure~\ref{fig:two possibilities LC factorization}), aspect ratio (2) $\sigma$, skewness factor (3) $\tau$,  principal center coordinates (4) $u$ and (5) $v$, 2D rotation angle (6) $\alpha$, projection center coordinates (7) $x_s$, (8) $y_s$ and (9) $z_s$; (10) $r$ and (11) $\theta$ are used together with $f$ and $x_s$, $y_s$, $z_s$ to represent projection plane $\pi$ in a minimum number of independent parameters. The key issue of the LC factorization of an arbitrary numerical pinhole projection matrix is to find all these 11 parameters.

It is well know from the classic textbooks~\cite{Faugeras:1993,Faugeras:2001,ZissermanHartley-5,Ma:2003:AnInvitation,Forsyth:2012} that, the Euclidean coordinates of the camera center (also called {\em projection center}), $\left(x_s, y_s, z_s\right)^T$, can be uniquely determined via the minus inverse of the left 3$\times$3 partition of the projection matrix multiplying its fourth column~\cite{Luong1996IJCV:Fmatrix, Faugeras:2001, ZissermanHartley-5}; and the normal direction of the projection plane(different from {\em principal plane}~\cite[p.158]{ZissermanHartley-5} which is the last row of projection matrix) are exactly the first three entries in the last row of the projection matrix, which therefore determines $r$, $\cos\theta$ and $\sin\theta$.

We will assume the $3\times 4$ pinhole camera matrix has been normalized into the form below before further processing:

\begin{equation}\label{eqn:pinhole camera matrix}
P_{\text{pinhole}}=
\left[
\begin{array}{cccc}
p_{11} &  p_{12} &  p_{13} & p_{14} \\
p_{21} &  p_{22} &  p_{23} & p_{24} \\
 r\cos\theta & r\sin\theta & \sqrt{1-r^2} & p_{34} \\
\end{array}
\right],\quad \text{where $0\leqslant r\leqslant 1$}
\end{equation}

In order to obtain the ``intrinsic parameters" $f$, $\sigma$, $\tau$, $u$ and $v$, we have the following result which is about the Kruppa matrix of an LC pinhole camera and is critical in intrinsic parameter determination~\cite{Faugeras:2001}~\cite[5~\texttildelow~43]{emergingTopics2005}:
\begin{proposition}[LC intrinsic parameters]\label{LC Kruppa proposition}
If $\boldsymbol M$ is the left 3$\times$3 submatrix of the pinhole camera projection matrix $\boldsymbol P$ in equation~\eqref{LC camera representation0}, then we have the following result:
\begin{equation}\label{LC Kruppa equation0}
{\boldsymbol M}\cdot {\boldsymbol M}^T = k\cdot \left[
\begin{array}{ccc}
 \left(\sigma ^2+\tau ^2\right) f^2+u^2 & \tau  f^2+u v & u \\
 \tau  f^2+u v & f^2+v^2 & v \\
 u & v & 1 \\
\end{array}
\right], \text{ $0\neq k \in {\mathbb{R}}$}
\end{equation}
\end{proposition}

By solving the equations in~\eqref{LC Kruppa equation0} we have $f$ fixed at some negative value (or positive, depending on one's preference for the two equivalent possibilities in Figure~\ref{fig:two possibilities LC factorization}), $\sigma$ (with sign to be determined), $\tau$, $u$ and $v$  for the different LC factorization options respectively. Then the cosine and sine values of the only left parameter $\alpha$ in equation~\eqref{LC camera representation0} can be obtained by solving a linear system of equations about $\cos\alpha$ and $\sin\alpha$. Denote $z^+=\sqrt{1-r^2}+1$, $z^-=\sqrt{1-r^2}-1$, then the coefficient matrix of this linear system, $A_{8\times 2}$, is as in equation~\eqref{eqn:pinhole A82}.

\begin{align}\label{eqn:pinhole A82}
A_{8\times 2}=\dfrac{f}{2}\left[
\begin{array}{cc}
-{z^-}(\sigma\cos{2\theta}+\tau\sin{2\theta})-{z^+}\sigma & {z^-}(\tau\cos{2\theta}-\sigma\sin{2\theta})+{z^+}\tau\\
{z^-}(\tau\cos{2\theta}-\sigma\sin{2\theta})-{z^+}\tau & {z^-}(\sigma\cos{2\theta}+\tau\sin{2\theta})-{z^+}\sigma\\
2r(\sigma\cos{\theta}+\tau\sin{\theta})&2r(\sigma\sin{\theta}-\tau\cos{\theta})\\
\color{red}A_{8\times 2}^*(4,1)& \color{red}A_{8\times 2}^*(4,2)\\
-{z^-} \sin{2\theta} & {z^-}\cos{2\theta}+{z^+}\\
{z^-}\cos{2\theta}-{z^+} & {z^-}\sin{2\theta}\\
2r\sin{\theta}&-2r\cos{\theta}\\
\color{red}A_{8\times 2}^*(8,1)& \color{red}A_{8\times 2}^*(8,2)\\
\end{array}
\right]
\end{align}
where:
\begin{align*}
{\color{red}A_{8\times 2}^*(4,1)} &={z^-}((\tau{x_s}+\sigma{y_s})\sin{2\theta}+(\sigma{x_s}-\tau{y_s})\cos{2\theta})+ {z^+}(\sigma{x_s}+\tau{y_s})-2r{z_s}(\sigma\cos{\theta}+\tau\sin{\theta})\nonumber\\
{\color{red}A_{8\times 2}^*(4,2)} &={z^-}((\sigma{x_s}-\tau{y_s})\sin{2\theta}-(\tau{x_s}+\sigma{y_s})\cos{2\theta})+{z^+}(\sigma{y_s}-\tau{x_s})+ 2r{z_s}(\tau\cos{\theta}-\sigma\sin{\theta})\nonumber\\
{\color{red}A_{8\times 2}^*(8,1)} &={z^-}({x_s}\sin{2\theta}-{y_s}\cos{2\theta})+{z^+}{y_s}-2r{z_s}\sin\theta\nonumber\\
{\color{red}A_{8\times 2}^*(8,2)} &={z^-}({x_s}\cos{2}\theta+{y_s}\sin{2}\theta)+{z^+}{x_s}-2r{z_s}\cos\theta\nonumber\\
\end{align*}
And a column vector $B_{8\times 3}$ which is to be subtracted from the first two rows of pinhole camera matrix:
\begin{equation}\label{eqn:sub8x1 vector pinhole}
B_{8\times 1}=\left(
\begin{array}{c}
 r u \cos \theta \\
 r u \sin \theta \\
 \sqrt{1-r^2} u \\
 -u \left(\sqrt{1-r^2} {z_s}+r {x_s} \cos \theta+r {y_s} \sin \theta\right) \\
 r v \cos \theta \\
 r v \sin \theta \\
 \sqrt{1-r^2} v \\
 -v \left(\sqrt{1-r^2} {z_s}+r {x_s} \cos \theta+r {y_s} \sin \theta\right) \\
\end{array}
\right)
\end{equation}

Let us rewritten the 8 entries in the first two rows of $P_{\text{pinhole}}$ as an $8\times 1$ vector:
\begin{equation}\label{eqn:alpha_map pinhole}
\left[
\begin{array}{cccc}
 p_{11} & p_{12} & p_{13} & p_{14} \\
 p_{21} & p_{22} & p_{23} & p_{24} \\
\end{array}
\right] \mapsto
T_{8\times 1}=\left(
\begin{array}{c}
 p_{11} \\
 p_{12} \\
 p_{13} \\
 p_{14} \\
 p_{21} \\
 p_{22} \\
 p_{23} \\
 p_{24} \\
\end{array}
\right)
\end{equation}

Then $X=(\cos\alpha, \sin\alpha)^T$ can be solved from the following $8\times 2$ linear system:
\begin{equation}\label{eqn:pinhole cossinalpha system}
A_{8\times 2}\cdot X = T_{8\times 1}-B_{8\times 1}
\end{equation}

The seven {\em left matrix factor}s (\circled{8}~\texttildelow~\circled{2}) can be concatenated together by matrix multiplication operation as the {\em three2two} conversion matrix which transforms 3D images generated by the central projection \circled{1} in the 3D world coordinate system into 2D images in the 2D image coordinate system.

{
\begin{algorithm}[!htp]
    \caption{LC factorization for pinhole cameras}
    \label{alg:LCFactorizationPinhole}
    \begin{algorithmic}[1]
        \Procedure{LCFactorizationXOYPnf}{$P_{3\times 4}$}\Comment{LC factorization}
        \State $(x_s,y_s,z_s,1)\gets$ null space of $P$~\eqref{eqn:pinhole camera matrix} in homogeneous form; \Comment{}
        \State $\left(r\cos\theta,r\sin\theta,\sqrt{1-r^2}\right)\gets$ the last row of $M$~\eqref{LC Kruppa equation0}; \Comment{}
        \State $\sigma,\tau, u, v, f\gets$Solve nonlinear systems per Kruppa matrix~\eqref{LC Kruppa equation0} \Comment{}
        \State $i \gets 0, n\gets$ solution number
        \For {$i=1$ \textbf{to} $n$}
        \State $\sin\alpha,\cos\alpha\gets$ solve linear system~\eqref{eqn:pinhole cossinalpha system};
        \If {$\cos^2\alpha+\sin^2\alpha=1$} \Comment{}
        \State calculate matrix factors\circled{1}-\circled{8} in equation~\eqref{LC camera representation0};
        \State {\textbf{return} \circled{1}-\circled{8} and current parameters}\Comment{comment this to get all possible cases}
        \EndIf
        \EndFor
       \State \textbf{return} matrix factors and all parameters\Comment{return all factorizations}
       \EndProcedure
   \end{algorithmic}
\end{algorithm}}

Note that in equation~\eqref{LC camera representation0} we assume that the $xoy$ axis plane is used as 2D image axis plane in LC factorization, which may also be $yoz$ or $xoz$ axis planes with the corresponding {\em cutting}\circled{3} and {\em reflection}\circled{2} factors changed accordingly. However, no matter which axis plane is used, the {\em three2two} results are only affected by the choice of oriented focal length $f$ in equation~\eqref{LC camera representation0} which can be positive or negative corresponding to the two possible cases of central projection \circled{1} as in Figure~\ref{fig:two possibilities LC factorization}. And there is only two possibilities of LC factorization when {\em three2two} is the only explicit {\em left matrix factor}.

The pseudo-code of a pinhole camera's LC factorization implementation is as in Algorithm~\ref{alg:LCFactorizationPinhole}.

It is also simple to obtain {\em two2three}, the inverse conversion matrix of {\em three2two}, which converts the 2D image coordinates into their 3D image counterparts.

LC factorization makes it possible to extract the inherent {\em central projection}s of calibrated pinhole cameras in the same 3D world coordinate system and convenient to implement image rectification and triangulation with 3D image information and central projection cameras.

For non-pinhole linear cameras, this will become critical in order to implement symmedian point triangulation which depends on the camera projection lines(principal rays)' 3D coordinate information.

\subsection{LC Factorization of Affine Cameras }\label{LC factorization affine}

It has been proved that an affine camera has the following matrix form~\cite{Faugeras:2001,ZissermanHartley-5,Forsyth:2012}:

\begin{equation}\label{eqn:affine camera matrix}
P_{\text{Affine}}=
\left[
\begin{array}{cccc}
a_{11} &  a_{12} &  a_{13} &  a_{14} \\
a_{21} &  a_{22} &  a_{23} &  a_{24} \\
 0 & 0 & 0 & 1 \\
\end{array}
\right]
\end{equation}

Different from pinhole cameras, the inherent {\em central projection}s of affine cameras has infinite $(s)$ projection centers, therefore the parameter {\em focal length} $f$ for pinhole camera becomes invalid. And the 3D image plane $(\pi)$ can almost be arbitrary except the infinite plane.

For the purpose of simplification, we fixed the image plane $(\pi)$ as the plane passing through the 3D coordinate origin with normal direction the same as $(s)$ in homogeneous form, i.e., algebraically, $(s)=(\pi)$.

Similar to the pinhole camera cases, let the LC factorization for an affine camera be:
\begin{align}\label{eqn:LC factorization affine}
P_{\text{Affine}}=\underbrace{\left[
\begin{array}{ccc}
 1 & 0 & u \\
 0 & 1 & v \\
 0 & 0 & 1 \\
\end{array}
\right]}_{\circled{6}}
\underbrace{\left[
\begin{array}{ccc}
 1 & \tau  & 0 \\
 0 & 1 & 0 \\
 0 & 0 & 1 \\
\end{array}
\right]}_{\circled{5}}
\underbrace{\left[
\begin{array}{ccc}
 \sigma  & 0 & 0 \\
 0 & \rho  & 0 \\
 0 & 0 & 1 \\
\end{array}
\right]}_{\circled{4}}
\underbrace{\left[
\begin{array}{ccc}
 \cos \alpha & \sin \alpha & 0 \\
 -\sin \alpha & \cos \alpha & 0 \\
 0 & 0 & 1 \\
\end{array}
\right]}_{\circled{3}}\nonumber\\
\underbrace{\left[
\begin{array}{cccc}
 1 & 0 & 0 & 0 \\
 0 & 1 & 0 & 0 \\
 0 & 0 & 0 & 1 \\
\end{array}
\right]}_{\circled{2}}\;\;
\underbrace{\left(I-2\dfrac{\left(S_R\right)\;\left(\pi_R\right)^T}{\left(S_R\right)^T\;\left(\pi_R\right)}\right)}_{\circled{1}}\;\;
\underbrace{\left(I-\dfrac{\left(S\right)\;\left(\pi\right)^T}{\left(S\right)^T\;\left(\pi\right)}\right)}_{\circled{0}}
\end{align}
where:
\begin{align*}
(S)=(\pi)= \left(r\cos\theta, r\sin\theta,\sqrt{1-r^2},0\right)^T, \text{are both the projection direction and the image plane}\\
\end{align*}
The factors \circled{1} defined as orthographic reflection, and \circled{0} orthographic parallel projection in~\eqref{eqn:LC factorization affine} have similar formula to the reflection~\circled{2} and projection~\circled{1} in~\eqref{LC camera representation0} with different stereohomology centers and hyperplanes as below.

Note that we assumed parallel projection~\circled{0} direction and image plane have the same homogeneous form, then $S=\pi=$ $\left(r\cos\theta,r\sin\theta,\sqrt{1-r^2},0\right)^T$. We number the matrix factors from $0$ because the matrix factor \circled{0} can actually be omitted without problem! Therefore, the orthographic parallel projection~\circled{0} is as:
\begin{equation}\label{eqn:parallel projection factor}
\circled{0}: \left(I - \dfrac{S\cdot \pi^T}{S^T\cdot \pi}\right)=\left[
\begin{array}{cccc}
 1 - r^2\cos ^2\theta & -r^2 \cos \theta \sin \theta & -r \sqrt{1-r^2} \cos \theta & 0 \\
 -r^2 \cos \theta \sin \theta & 1- r^2\sin ^2\theta & -r \sqrt{1-r^2} \sin \theta & 0 \\
 -r \sqrt{1-r^2} \cos \theta & -r \sqrt{1-r^2} \sin \theta & r^2 & 0 \\
 0 & 0 & 0 & 1 \\
\end{array}
\right]
\end{equation}

The reflection \circled{1} which transforms $\pi=\left(r\cos\theta,r\sin\theta,\sqrt{1-r^2},0\right)^T$ into axis plane $xoy$ is the orthographic reflection with mirror plane $\pi_{\text{R}}$ $=\left(r\cos\theta,r\sin\theta,\sqrt{1-r^2}+1,0\right)^T$, the normal direction of which can be easily obtained: $S_{\text{R}}$ $=\left(r\cos\theta,r\sin\theta,\sqrt{1-r^2}+1,0\right)^T$. So that we have:
\begin{equation}\label{eqn:ortho-reflection factor affine}
\circled{1}: {\left(I - 2\dfrac{S_\text{R}\cdot \pi_\text{R}^T}{S_\text{R}^T\cdot \pi_\text{R}}\right)}=\left[
\begin{array}{cccc}
 \left(\sqrt{1-r^2}-1\right) \cos ^2\theta+1 & \left(\sqrt{1-r^2}-1\right) \cos \theta \sin \theta & -r \cos \theta & 0 \\
 \left(\sqrt{1-r^2}-1\right) \cos \theta \sin \theta & \left(\sqrt{1-r^2}-1\right) \sin ^2\theta+1 & -r \sin \theta & 0 \\
 -r \cos \theta & -r \sin \theta & -\sqrt{1-r^2} & 0 \\
 0 & 0 & 0 & 1 \\
\end{array}
\right]
\end{equation}

There are only 8 independent parameters, $u$, $v$, $\sigma$, $\tau$, $\rho$, $\alpha$, $\theta$ and $r$, in total. Therefore, for any given {\em affine camera} matrix in the form~\eqref{eqn:affine camera matrix}, implementing LC factorization is actually equivalent to determining the 8 parameters.

Similar to the Kruppa's matrix for pinhole cameras in Lemma~\ref{LC Kruppa proposition}, it is easy to prove for $P_{\text{Affine}}$ in~\eqref{eqn:LC factorization affine}:
\begin{equation}\label{eqn:Kruppa matrix affine}
P_{\text{Affine}}\cdot P_{\text{Affine}}^T =
\left[
\begin{array}{ccc}
 u^2+\sigma ^2+\rho ^2 \tau ^2 & \tau  \rho ^2+u v & u \\
 \tau  \rho ^2+u v & v^2+\rho ^2 & v \\
 u & v & 1 \\
\end{array}
\right]
\end{equation}
Such that 5 of the 8 parameters $u$, $v$ and $\tau$ can be determined easily, while $\sigma$ and $\rho$ can also be determined up to a plus or minus sign. If we fix $sigma$ as a positive value, then the sign of $\rho$ can be determined per the calculation of sine and cosine values of $\alpha$.

Since $\left(r\cos\theta,r\sin\theta,\sqrt{1-r^2}\right)^T$ is the null space of the left $3\times 3$ partition of $P_{\text{Affine}}$, another 2 parameters $r$ and $\theta$ can be determined easily this way. Then there is only $\alpha$ left to be determined.

Denote the left-upper $2\times 3$ partition of $P_{\text{Affine}}$ as $B_{2\times 3}$, then reshape it into $B_{6\times 1}$ vector:
\begin{equation}\label{eqn:alpha_map affine}
\left[
\begin{array}{ccc}
 a_{11} & a_{12} & a_{13} \\
 a_{21} & a_{22} & a_{23} \\
\end{array}
\right] \mapsto
B_{6\times 1}=
\left(
\begin{array}{c}
a_{11} \\
a_{12} \\
a_{13} \\
a_{21} \\
a_{22} \\
a_{23} \\
\end{array}
\right)_{6\times 1}
\end{equation}

Let $X_{2\times 1}=\left(\cos\alpha,\sin\alpha\right)^T$, denote $z^+ =\sqrt{1-r^2}+1, z^-=\sqrt{1-r^2}-1$, and:
\noindent\begin{align}A_{6\times 2}=\dfrac{1}{2}
\left[
\begin{array}{cc}
-{z^-}\sigma-{z^+}(\sigma\cos{2\theta}+\rho\tau\sin{2\theta})&{z^-}\rho\tau+{z^+}(\rho\tau\cos{2\theta}-\sigma\sin{2\theta})\\
{z^+}(\rho\tau-\sigma\sin{2\theta})-{z^-}\rho\tau\cos{2\theta}&{z^+}(\sigma\cos{2\theta}+\rho\tau\sin{2\theta})-{z^-}\sigma\\
2r(\sigma\cos{\theta}+\rho\tau\sin{\theta})&2r(\sigma\sin{\theta}-\rho\tau\cos{\theta})\\
-{z^+}\rho\sin{2\theta}&\rho\left({z^-}+{z^+}\cos{2\theta}\right)\\
\rho\left({z^+}\cos{2\theta}-{z^-}\right)&{z^+}\rho\sin{2\theta}\\
2r\rho\sin{\theta}&-2r\rho\cos{\theta}\\
\end{array}
\right]_{6\times 2}\end{align}
then $\cos\alpha$ and $\sin\alpha$ can be solved from the $6\times 2$ linear system:
\begin{equation}\label{eqn:affine cossinalpha system}
A_{6\times 2} \cdot X_{2\times 1} = B_{6\times 1}
\end{equation}
The $\rho$ which satisfies $\cos^2\alpha+\sin^2\alpha=1$ is the desired $\rho$.

The pseudo-code of the affine camera LC factorization is as in Algorithm~\ref{alg:LCFactorizationAffine}.
{
\begin{algorithm}[!htp]
    \caption{LC factorization for affine cameras}
    \label{alg:LCFactorizationAffine}
    \begin{algorithmic}[1]
        \Procedure{LCFactorizationXOYAfp}{$P_{3\times 4}$}\Comment{LC factorization}
        \State $\left(r\cos\theta,r\sin\theta,\sqrt{1-r^2},0\right)\gets$ null space of $P$~\eqref{eqn:affine camera matrix} in homogeneous form; \Comment{}
        \State $\sigma,\rho, \tau, u, v\gets$Solve nonlinear systems per Kruppa matrix~\eqref{eqn:Kruppa matrix affine} \Comment{}
        \State $i \gets 0, n\gets$ solution number
        \For {$i=1$ \textbf{to} $n$}
        \State $\sin\alpha,\cos\alpha\gets$ solve linear system~\eqref{eqn:affine cossinalpha system};
        \If {$\cos^2\alpha+\sin^2\alpha=1$} \Comment{}
        \State calculate matrix factors\circled{0}-\circled{6} in equation~\eqref{eqn:LC factorization affine};
        \State {\textbf{return} \circled{0}-\circled{6} and current parameters}\Comment{comment this to get all possible cases}
        \EndIf
        \EndFor 
       \State \textbf{return} matrix factors and all parameters\Comment{return all factorizations}
       \EndProcedure
   \end{algorithmic}
\end{algorithm}}

There is another kind of linear cameras which are neither pinhole nor affine cameras~\eqref{eqn:affine camera matrix}. The left $3\times 3$ partitions of them have a matrix rank of 2 and their third rows are not zero. Since by only left row elementary transformations with at most two additional individual parameters, any other non-pinhole camera $3\times 4$ projection matrices can be transformed into such affine cameras as in~\eqref{eqn:affine camera matrix}, their LC factorization can be easily implemented the same way as those of affine cameras after simple row elementary transformations~\eqref{eqn:elementary row transforms}. The implementation details will be omitted here.
\begin{align}\label{eqn:elementary row transforms}
\left[
\begin{array}{ccc}
 1 & 0 & 0 \\
 0 & 1 & 0 \\
 \lambda _1 & \lambda _2 & 1 \\
\end{array}
\right], \quad
\left[
\begin{array}{ccc}
 1 & 0 & 0 \\
 0 & 0 & 1 \\
 0 & 1 & 0 \\
\end{array}
\right](\text{optional}),\quad
\left[
\begin{array}{ccc}
 0 & 0 & 1 \\
 0 & 1 & 0 \\
 1 & 0 & 0 \\
\end{array}
\right](\text{optional})
\end{align}

LC factorizations are generally more complicate than conventional factorization methods. But they make it possible to extract the $4\times 4$ {\em projection} matrices from $3\times 4$ linear camera matrices; we can easily obtain the matrix factors of the conventional decomposition from LC factorization results; we can extract the projection line(principal ray) information easily and the 3D image coordinate information when necessary.

There are two immediately comparison results between LC factorization and the conventional method. First, principal rays (which we call projection lines) of both pinhole and affine cameras and sometimes 3D images can be extracted easily, which makes it possible that symmedian point triangulation is extended to affine camera cases without difficulty; second, $KRt$ can be derived from LC factors and the physical meanings of LC factors and some explicit parameters are made clearer, the advantage of which over $KRt$ will be depicted in a separate work.

\section{Applications of LC Factorization in Triangulation}\label{symmedian point triangulation section}

We show how LC factorization can be use to replace conventional $KRt$ decomposition and extend the applications of camera decomposition to beyond the capability of $KRt$ decomposition.

\subsection{Generalization of mid-point triangulation to multiple view cases}
\noindent The object of two-view midpoint triangulation method is to find the midpoint of the common perpendicular segment of the two projection lines(principal rays). Such concept as midpoint can be generalized into multiple view cases if we use its two-view equivalent: a point with the minimum sum of squared distances to all the projection lines(principal rays) of different views. This generalization can be called ``symmedian point" triangulation.

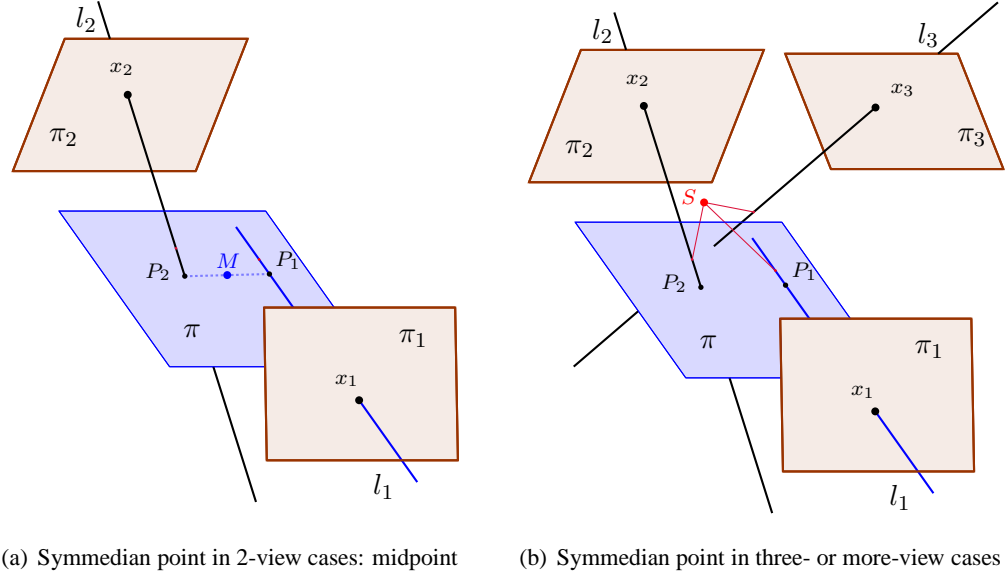
\begin{figure}[!hpt]
\begin{center}
\subfigure[Symmedian point in 2-view cases: midpoint]{\label{fig:(a)midpoint}
\newrgbcolor{zzttqq}{0.6 0.2 0.}
\newrgbcolor{xdxdff}{0.490196078431 0.490196078431 1.}
\newrgbcolor{dcrutc}{0.862745098039 0.078431372549 0.235294117647}
\psset{xunit=0.75cm,yunit=0.75cm,algebraic=true,dimen=middle,dotstyle=o,dotsize=3pt 0,linewidth=0.8pt,arrowsize=3pt 2,arrowinset=0.25}
\begin{pspicture*}(1.23041106647,-4.08426207511)(9.93779534756,5.60176193786)
\pspolygon[linecolor=zzttqq,fillcolor=zzttqq,fillstyle=solid,opacity=0.1](2.4612,4.6176)(1.5416,2.2702)(4.7844,2.2702)(5.704,4.6176)
\pspolygon[linewidth=0.4pt,linecolor=blue,fillcolor=blue,fillstyle=solid,opacity=0.15](6.01315357143,-1.2)(4.32,-1.2)(2.3596107571,1.56173637805)(6.02323642004,1.56173637805)(7.23815234432,-0.1498)(5.9944,-0.1498)
\pspolygon[linecolor=zzttqq,fillcolor=zzttqq,fillstyle=solid,opacity=0.1](6.0428,-2.8602)(9.4308,-2.8602)(9.3824,-0.1498)(5.9944,-0.1498)
\psline[linecolor=zzttqq](2.4612,4.6176)(1.5416,2.2702)
\psline[linecolor=zzttqq](1.5416,2.2702)(4.7844,2.2702)
\psline[linecolor=zzttqq](4.7844,2.2702)(5.704,4.6176)
\psline[linecolor=zzttqq](5.704,4.6176)(2.4612,4.6176)
\psline(3.05330342045,5.282126571)(3.2626974363,4.6176)
\psline(3.57618665292,3.62272002062)(4.59018665292,0.404720020625)
\psline[linestyle=dashed,dash=1pt 1pt,linecolor=xdxdff](4.59018665292,0.404720020625)(6.09018665292,0.444720020625)
\psline[linecolor=blue](8.71030698232,-3.24642532451)(7.68052077522,-1.79569402711)
\psline[linecolor=blue](5.49738159068,1.27984565264)(6.51220037674,-0.1498)
\psline(5.85178673451,-3.59905617519)(5.09583802051,-1.2)
\rput[tl](2.66073067461,5.17517538806){$l_2$}
\rput[tl](7.93032923088,-3.10562234323){$l_1$}
\rput[tl](8.38200910713,-0.546103044464){$\pi_1$}
\rput[tl](2.18395747189,3.01714931264){$\pi_2$}
\rput[tl](4.56782348545,-0.44572973863){$\pi $}
\psline[linewidth=0.4pt,linecolor=blue](6.01315357143,-1.2)(4.32,-1.2)
\psline[linewidth=0.4pt,linecolor=blue](4.32,-1.2)(2.3596107571,1.56173637805)
\psline[linewidth=0.4pt,linecolor=blue](2.3596107571,1.56173637805)(6.02323642004,1.56173637805)
\psline[linewidth=0.4pt,linecolor=blue](6.02323642004,1.56173637805)(7.23815234432,-0.1498)
\psline[linewidth=0.4pt,linecolor=blue](7.23815234432,-0.1498)(5.9944,-0.1498)
\psline[linewidth=0.4pt,linecolor=blue](5.9944,-0.1498)(6.01315357143,-1.2)
\psline[linecolor=zzttqq](6.0428,-2.8602)(9.4308,-2.8602)
\psline[linecolor=zzttqq](9.4308,-2.8602)(9.3824,-0.1498)
\psline[linecolor=zzttqq](9.3824,-0.1498)(5.9944,-0.1498)
\psline[linecolor=zzttqq](5.9944,-0.1498)(6.0428,-2.8602)
\begin{scriptsize}
\psdots[dotstyle=*](3.57618665292,3.62272002062)
\rput[bl](3.26297050961,3.97069571806){$x_2$}
\psdots[dotsize=2pt 0,dotstyle=*](4.59018665292,0.404720020625)
\rput[bl](3.89030367107,0.332163381582){$P_2$}
\psdots[dotsize=2pt 0,dotstyle=*](6.09018665292,0.444720020625)
\rput[bl](6.19888970525,0.532909993249){$P_1$}
\psdots[dotstyle=*,linecolor=blue](5.34018665292,0.424720020625)
\rput[bl](5.14496999399,0.558003319708){\blue{$M$}}
\psdots[dotstyle=*](7.68052077522,-1.79569402711)
\rput[bl](7.2528094165,-1.5498361028){$x_1$}
\psdots[dotsize=1pt 0,dotstyle=*,linecolor=dcrutc](4.43439969502,0.899120839684)
\psdots[dotsize=1pt 0,dotstyle=*,linecolor=red](5.90805480325,0.701301797259)
\end{scriptsize}
\end{pspicture*}
}
\subfigure[Symmedian point in three- or more-view cases]{\label{fig:(b)symmedian point}
\newrgbcolor{zzttqq}{0.6 0.2 0.}
\newrgbcolor{dcrutc}{0.862745098039 0.078431372549 0.235294117647}
\psset{xunit=0.75cm,yunit=0.75cm,algebraic=true,dimen=middle,dotstyle=o,dotsize=3pt 0,linewidth=0.8pt,arrowsize=3pt 2,arrowinset=0.25}
\begin{pspicture*}(1.43115767814,-3.88351546344)(10.1134486328,5.50138863202)
\pspolygon[linecolor=zzttqq,fillcolor=zzttqq,fillstyle=solid,opacity=0.1](2.4612,4.6176)(1.5416,2.2702)(4.7844,2.2702)(5.704,4.6176)
\pspolygon[linecolor=zzttqq,fillcolor=zzttqq,fillstyle=solid,opacity=0.1](6.89283554007,2.50405463415)(9.95086109175,2.50405463415)(9.14138373984,4.55023349593)(6.08335818815,4.55023349593)
\pspolygon[linewidth=0.4pt,linecolor=blue,fillcolor=blue,fillstyle=solid,opacity=0.15](6.01315357143,-1.2)(4.32,-1.2)(2.3596107571,1.56173637805)(6.02323642004,1.56173637805)(7.23815234432,-0.1498)(5.9944,-0.1498)
\pspolygon[linecolor=zzttqq,fillcolor=zzttqq,fillstyle=solid,opacity=0.1](6.0428,-2.8602)(9.4308,-2.8602)(9.3824,-0.1498)(5.9944,-0.1498)
\psline[linecolor=zzttqq](2.4612,4.6176)(1.5416,2.2702)
\psline[linecolor=zzttqq](1.5416,2.2702)(4.7844,2.2702)
\psline[linecolor=zzttqq](4.7844,2.2702)(5.704,4.6176)
\psline[linecolor=zzttqq](5.704,4.6176)(2.4612,4.6176)
\psline(3.05330342045,5.282126571)(3.2626974363,4.6176)
\psline(3.57618665292,3.62272002062)(4.59018665292,0.404720020625)
\psline[linecolor=blue](8.71030698232,-3.24642532451)(7.68052077522,-1.79569402711)
\psline[linecolor=blue](5.49738159068,1.27984565264)(6.51220037674,-0.1498)
\psline(5.85178673451,-3.59905617519)(5.09583802051,-1.2)
\psline[linecolor=zzttqq](6.89283554007,2.50405463415)(9.95086109175,2.50405463415)
\psline[linecolor=zzttqq](9.95086109175,2.50405463415)(9.14138373984,4.55023349593)
\psline[linecolor=zzttqq](9.14138373984,4.55023349593)(6.08335818815,4.55023349593)
\psline[linecolor=zzttqq](6.08335818815,4.55023349593)(6.89283554007,2.50405463415)
\psline(2.33401267567,-0.999878086351)(3.47936042943,-0.0157326963114)
\psline(4.81875675003,1.13514982825)(7.68504309891,3.59801971865)
\psline(8.79322886288,4.55023349593)(9.85384433433,5.46157227789)
\rput[tl](2.66073067461,5.17517538806){$l_2$}
\rput[tl](7.93032923088,-3.10562234323){$l_1$}
\rput[tl](8.43219576005,5.09989540869){$l_3$}
\rput[tl](8.38200910713,-0.546103044464){$\pi_1$}
\rput[tl](2.18395747189,3.01714931264){$\pi_2$}
\rput[tl](9.13480890089,3.2178959243){$\pi_3$}
\rput[tl](4.56782348545,-0.44572973863){$\pi $}
\psline[linewidth=0.4pt,linecolor=dcrutc](4.64310346482,1.91304294846)(4.43439969502,0.899120839684)
\psline[linewidth=0.4pt,linecolor=dcrutc](4.64310346482,1.91304294846)(5.52575361752,1.74264017834)
\psline[linewidth=0.4pt,linecolor=dcrutc](4.64310346482,1.91304294846)(5.90805480325,0.701301797259)
\psline[linewidth=0.4pt,linecolor=blue](6.01315357143,-1.2)(4.32,-1.2)
\psline[linewidth=0.4pt,linecolor=blue](4.32,-1.2)(2.3596107571,1.56173637805)
\psline[linewidth=0.4pt,linecolor=blue](2.3596107571,1.56173637805)(6.02323642004,1.56173637805)
\psline[linewidth=0.4pt,linecolor=blue](6.02323642004,1.56173637805)(7.23815234432,-0.1498)
\psline[linewidth=0.4pt,linecolor=blue](7.23815234432,-0.1498)(5.9944,-0.1498)
\psline[linewidth=0.4pt,linecolor=blue](5.9944,-0.1498)(6.01315357143,-1.2)
\psline[linecolor=zzttqq](6.0428,-2.8602)(9.4308,-2.8602)
\psline[linecolor=zzttqq](9.4308,-2.8602)(9.3824,-0.1498)
\psline[linecolor=zzttqq](9.3824,-0.1498)(5.9944,-0.1498)
\psline[linecolor=zzttqq](5.9944,-0.1498)(6.0428,-2.8602)
\begin{scriptsize}
\psdots[dotstyle=*](3.57618665292,3.62272002062)
\rput[bl](3.26297050961,3.97069571806){$x_2$}
\psdots[dotsize=2pt 0,dotstyle=*](4.59018665292,0.404720020625)
\rput[bl](3.89030367107,0.332163381582){$P_2$}
\psdots[dotsize=2pt 0,dotstyle=*](6.09018665292,0.444720020625)
\rput[bl](6.19888970525,0.532909993249){$P_1$}
\psdots[dotstyle=*](7.68052077522,-1.79569402711)
\rput[bl](7.2528094165,-1.5498361028){$x_1$}
\psdots[dotstyle=*](7.68504309891,3.59801971865)
\rput[bl](7.95542255734,3.79504243285){$x_3$}
\psdots[dotstyle=*,linecolor=red](4.64310346482,1.91304294846)
\rput[bl](4.24161024149,1.88794962201){\red{$S$}}
\psdots[dotsize=1pt 0,dotstyle=*,linecolor=dcrutc](4.43439969502,0.899120839684)
\psdots[dotsize=1pt 0,dotstyle=*,linecolor=red](5.90805480325,0.701301797259)
\psdots[dotsize=1pt 0,dotstyle=*,linecolor=dcrutc](5.52575361752,1.74264017834)
\end{scriptsize}
\end{pspicture*}
}
\end{center}
\caption{Demonstration of the generalization from mid-point to symmedian-point triangulation}
\label{fig:symmedian point}
\end{figure}

Such an idea initially appeared in~\cite{sturm:midpoint} by Sturm et al (2006) and a simple implementation for pinhole camera cases was given in~\cite[pp.305~\texttildelow~307]{Szeliski:ComputerVision2011} by Szeliski (2010). Such method is linear, suboptimal and efficient, the results obtained in closed form can be used as high quality initial values for further $L2$ optimal improvement~\cite{iterativeSolver2014}. A detailed report on symmedian point triangulation has been given based on the Oxford VGG data sets~\cite{iterativeSolver2014}. No other methods asserts L2 optimal triangulation for these data sets with such efficiency. For three or more view triangulation cases, the iterative method initialized by symmedian points~\cite{iterativeSolver2014} might be the {\em gold standard algorithm} in this means.

However, we have found no implementation of such triangulation for affine cameras cases due to the limitation of conventional camera model and its decomposition. By using LC factorization such triangulation can be implemented for both pinhole and affine cameras. We will depict the details in this section.

In 3D Euclidean space, a line $l_i$ is defined by a fixed point $X_i$ and its direction $W_i$ as:
\begin{equation}
{l_i} \triangleq \left\langle {{X_i},{W_i}} \right\rangle
\end{equation}
where $W_i$ is a unit vector.  Define the $3\times 3$ Euclidean projection matrix $P_i$ as:
\begin{equation}
{P_i} \triangleq I - \frac{{{W_i} \cdot W_i^T}}{{W_i^T \cdot {W_i}}} = I - {W_i} \cdot W_i^T
\end{equation}
such that the distance $d_i$ between a point $X$ and line $l_i$ $= \left\langle {{X_i},{W_i}} \right\rangle $ satisfies:
\begin{equation}
{d_i^2} = \left\| {{P_i}\left( {X - {X_i}} \right)} \right\|_2^2
\end{equation}
 then the symmedian point $X^*$, the sum of squared distances of which to all the $l_i$ lines is at minimum, is given by the solution to the following $3\times 3$ linear system:
\begin{equation}\label{symmedian point method linear equations}
\left( {\sum\limits_{i = 1}^n {{P_i}} } \right) \cdot X = \sum\limits_{i = 1}^n {\left( {{P_i} \cdot {X_i}} \right)} \end{equation}
So generally we can have the following result when the linear equations have a unique solution:
\begin{equation}\label{symmedian point solution closed form}
X^* = {\left( {\sum\limits_{i = 1}^n {{P_i}} } \right)^\dag } \cdot \sum\limits_{i = 1}^n {\left( {{P_i} \cdot {X_i}} \right)}
\end{equation}
or by using $QR$ decomposition and back-substitution solution:
\begin{equation}\label{symmedian point solution numeric}
X^* = \left( {\sum\limits_{i = 1}^n {{P_i}} } \right)\backslash \sum\limits_{i = 1}^n {\left( {{P_i} \cdot {X_i}} \right)}
\end{equation}
Note the above symmedian point triangulation algorithm fails when the coefficient matrix of system~\eqref{symmedian point method linear equations} is singular which means all $l_i$ are parallel to each other. Such cases rarely occur in real practices and usually mean large calibration error or miss matching, so it is legitimate to omit such data if there are any.
\subsection{Find projection lines(principal rays) via camera factorizations}

The key to implement the above triangulation is to find all the $W_i$ and $X_i$ from the 2D images $U_i$ and the calibrated cameras $P_i$. An implementation for pinhole cameras~\cite{Szeliski:ComputerVision2011} is dependent on $KRt$ decomposition, where $S_i$ is used as $X_i$ and $W_i = R_i^T K_i^{-1}U_i$. This is simple, efficient and probably the best approach to determine the project line of a pinhole camera.

While for affine camera cases, since $S_i$ in such cases are infinite points which can only be represented by homogeneous coordinates, the conventional decomposition methods do not seem to deal with it easily: as an infinite point $S_i$ can only be used as $W_i$, direction of the line, and there is still a lack of one ordinary point so as to determine the projection lines (rays). The triangulation implementation requires ``LC factorization" of pinhole cameras such that $3D$ image coordinates, $X_i$  and $W_i$ used can be obtained efficiently.

In order to determine the projection lines(principal rays), we use 3D image points as $X_i$'s since they are always ordinary points, and use the affine cameras' projection directions $S_i$ or the unit direction vector from 3D image points to the pinhole optical centers $S_i$ as $W_i$'s.  Conversion matrices {\em three2two} and {\em two2three} can be obtained from the left matrix factors of LC factorization for pinhole and affine cameras.  A key feature of LC factorization is that the conversions between 2D images and 3D images have thus been made invertible.

For example, per equation~\eqref{LC factorization pinhole} for pinhole cameras, the {\em two2three} conversion is as follows:
\begin{align}
k\cdot\underbrace {\left(
\begin{array}{c}
 x_i \\
 y_i \\
 z_i \\
 1 \\
\end{array}
\right)}_{\text{3D image}}
= \underbrace {\left( {I - 2\frac{{\left( {{S_R}} \right){{\left( {{\pi _R}} \right)}^ T }}}{{{{\left( {{S_R}} \right)}^ T }\left( {{\pi _R}} \right)}}} \right)}_{\circled{a}}
\underbrace {\left[ {\begin{array}{*{20}{c}}
  1&0&0 \\
  0&1&0 \\
  0&0&0 \\
  0&0&1
\end{array}} \right]}_{\circled{b}}
\underbrace {\left[ {\begin{array}{*{20}{c}}
  1&0&{  u_s} \\
  0&1&{  v_s} \\
  0&0&1
\end{array}} \right]}_{\circled{c}}
\underbrace {\left[ {\begin{array}{*{20}{c}}
  {\cos\alpha}&{-\sin\alpha}&0 \\
  {\sin\alpha}&{\cos\alpha }&0 \\
  0&0&1
\end{array}} \right]}_{\circled{d}}\nonumber\hfill\\
\underbrace {\left[ {\begin{array}{*{20}{c}}
  \frac{1}{\sigma}&0&{  0} \\
  0&1&{  0} \\
  0&0&1
\end{array}} \right]}_{\circled{e}}
\underbrace {\left[ {\begin{array}{*{20}{c}}
  1&-\tau&{  0} \\
  0&1&{  0} \\
  0&0&1
\end{array}} \right]}_{\circled{f}}
\underbrace {\left[ {\begin{array}{*{20}{c}}
  1&0&{-u} \\
  0&1&{-v} \\
  0&0&1
\end{array}} \right]}_{\circled{g}}
\underbrace {\left(
\begin{array}{c}
 u_i \\
 v_i \\
 1 \\
\end{array}
\right)}_{\text{2D image}}
\end{align}

For affine camera in equation~\eqref{eqn:LC factorization affine}, the conversion from 2D image to 3D image is as:
\begin{align}
k\cdot\underbrace {\left(
\begin{array}{c}
 x_i \\
 y_i \\
 z_i \\
 1 \\
\end{array}
\right)}_{\text{3D image}}=
\underbrace{\left(I-2\dfrac{\left(S_R\right)\;\left(\pi_R\right)^T}{\left(S_R\right)^T\;\left(\pi_R\right)}\right)}_{\circled{A}}
\underbrace{\left[
\begin{array}{cccc}
 1 & 0 & 0\\
 0 & 1 & 0\\
 0 & 0 & 0\\
 0 & 0 & 1
\end{array}
\right]}_{\circled{B}}
\underbrace{\left[
\begin{array}{ccc}
 \cos\alpha &-\sin\alpha & 0 \\
 \sin\alpha & \cos\alpha & 0 \\
 0 & 0 & 1 \\
\end{array}
\right]}_{\circled{C}}
\underbrace{\left[
\begin{array}{ccc}
 \frac{1}\sigma  & 0 & 0 \\
 0 & \frac{1}\rho  & 0 \\
 0 & 0 & 1 \\
\end{array}
\right]}_{\circled{D}}\nonumber\hfill\\
\underbrace{\left[
\begin{array}{ccc}
 1 & -\tau  & 0 \\
 0 & 1 & 0 \\
 0 & 0 & 1 \\
\end{array}
\right]}_{\circled{E}}
\underbrace{\left[
\begin{array}{ccc}
 1 & 0 & -u \\
 0 & 1 & -v \\
 0 & 0 & 1 \\
\end{array}
\right]}_{\circled{F}}
\underbrace {\left(
\begin{array}{c}
 u_i \\
 v_i \\
 1 \\
\end{array}
\right)}_{\text{2D image}}
\end{align}
Once 3D images are obtained from the 2D images measured, with their homogenous coordinates converted into Euclidean ones, the lines joining the 3D images obtained with the centers $S_i$, whether infinite or not, determine the desired projection lines(principal rays) in symmedian point triangulation. This solves the multiple view symmedian point triangulation for both pinhole and affine cameras.

Note that the method given in~\cite{Szeliski:ComputerVision2011} for pinhole cameras is not only equivalent to but also slightly simpler than the LC factorization based method here. Other applications of LC factorization on pinhole cameras and its comparisons with $KRt$ decomposition will be depicted in separate works.

\section{Appendices}\label{examples section}


\subsection{A conventional decomposition technique}

Though the intrinsic parameters can be obtained through Kruppa's matrix by solving nonlinear systems~\eqref{LC Kruppa equation0} and~\eqref{eqn:Kruppa matrix affine}, a more elegant and immediate implementation of $KRt$ decomposition is usually based on a very smart $RQ$ decomposition technique via $QR$ decomposition and Householder transformation~\cite{ZissermanHartley-5,VGG2012}.

For $A_{3\times 3}=\left(a_{i,j}\right)_{3\times 3}$, denote
\begin{align}
I_r(3)=\left[
\begin{array}{ccc}
 0 & 0 & 1 \\
 0 & 1 & 0 \\
 1 & 0 & 0 \\
\end{array}
\right], \text{ and }
A_{3\times 3}^{Trr}=I_r(3)\;A_{3\times 3}^T
\;I_r(3)
=I_r(3)\; \left[
\begin{array}{ccc}
 a_{1,1} & a_{1,2} & a_{1,3} \\
 a_{2,1} & a_{2,2} & a_{2,3} \\
 a_{3,1} & a_{3,2} & a_{3,3} \\
\end{array}
\right]^T \;I_r(3)
\end{align}
Suppose the $QR$ decomposition of $A_{3\times 3}^{Trr}$ is as below:
\begin{align}
A_{3\times 3}^{Trr}=
\left[
\begin{array}{ccc}
 a_{3,3} & a_{2,3} & a_{1,3} \\
 a_{3,2} & a_{2,2} & a_{1,2} \\
 a_{3,1} & a_{2,1} & a_{1,1} \\
\end{array}
\right]=Q R=
\left[
\begin{array}{ccc}
 q_{1,1} & q_{1,2} & q_{1,3} \\
 q_{2,1} & q_{2,2} & q_{2,3} \\
 q_{3,1} & q_{3,2} & q_{3,3} \\
\end{array}
\right]
\left[
\begin{array}{ccc}
 r_{1,1} & r_{1,2} & r_{1,3} \\
 0 & r_{2,2} & r_{2,3} \\
 0 & 0 & r_{3,3} \\
\end{array}
\right]
\end{align}
Then:
\begin{flalign}
A_{3\times 3}&=I_r(3)\;
\left(A_{3\times 3}^{Trr}\right)^T
\;I_r(3) =
I_r(3)
R^T
Q^T
I_r(3) = I_r(3)\;
R^T\; I_r(3)\; I_r(3)\;
Q^T\; I_r(3) \hphantom{AAAAAAA}\nonumber\hfill\\
&=\left(
I_r(3)\;
\left[
\begin{array}{ccc}
 r_{1,1} & 0 & 0 \\
 r_{1,2} & r_{2,2} & 0 \\
 r_{1,3} & r_{2,3} & r_{3,3} \\
\end{array}
\right]
I_r(3)
\right)
\left(
I_r(3)\left[
\begin{array}{ccc}
 q_{1,1} & q_{2,1} & q_{3,1} \\
 q_{1,2} & q_{2,2} & q_{3,2} \\
 q_{1,3} & q_{2,3} & q_{3,3} \\
\end{array}
\right]\;
I_r(3)\right)\hphantom{AAAAAAAAAAAAAAAAAAAA}\nonumber\hfill\\
&=\left[
\begin{array}{ccc}
 r_{3,3} & r_{2,3} & r_{1,3} \\
 0 & r_{2,2} & r_{1,2} \\
 0 & 0 & r_{1,1} \\
\end{array}
\right] \left[
\begin{array}{ccc}
 q_{3,3} & q_{2,3} & q_{1,3} \\
 q_{3,2} & q_{2,2} & q_{1,2} \\
 q_{3,1} & q_{2,1} & q_{1,1} \\
\end{array}
\right]\hphantom{AAAAAAAAAAAAAAAAAAAA}
\end{flalign}

In LC factorization Algorithms~\ref{alg:LCFactorizationPinhole} and~\ref{alg:LCFactorizationAffine},  more solutions than right ``intrinsic parameters" are first obtained by solving nonlinear systems based on the Kruppa like matrices~\eqref{LC Kruppa equation0}(in proposition~\ref{LC Kruppa proposition}) and~\eqref{eqn:Kruppa matrix affine}, then right parameters can be determined by solving linear systems of $\cos\alpha$ and $\sin\alpha$ and checking whether they can meet the $\cos^2\alpha+\sin^2\alpha=1$ criterion. The $RQ$ decomposition gives the right ``intrinsic parameters" immediately without the need of further verification and is suggested to use as an efficient alternative to the standard method.

\subsection{Critical matrices for other implementations of LC factorization}

When factoring a pinhole camera into its LC factors, the $\left(\pi_R\right)$ used in~\eqref{LC camera representation0} can also be $\left(r\cos\theta,r\sin\theta,\sqrt{1-r^2}- 1,d\right)^T$; in order to implement such LC factorization, differences take place in computation are only {\em reflection}, $u_s$, $v_s$ and the $A_{8\times 2}$ matrix. We only give the new $A_{8\times 2}$ here. Denote $z^+=\sqrt{1-r^2}+1$ and $z^-=\sqrt{1-r^2}-1$, then:
\begin{align}
A_{8\times 2}=\dfrac{f}{2}\left[
\begin{array}{ccc}
(\sigma\cos2\theta+\tau\sin2\theta)z^++z^-\sigma&&z^+(\sigma\sin2\theta-\tau\cos2\theta)-\tau z^-\\
z^+(\sigma\sin{2\theta}-\tau\cos{2\theta})+z^-\tau &&-z^+(\sigma\cos{2\theta}+\tau\sin{2\theta})+z^-\sigma\\
-2r(\sigma\cos\theta+\tau\sin\theta) &&2r(\tau\cos\theta-\sigma\sin\theta)\\
\color{red}A_{8\times 2}^*(4,1) && \color{red}A_{8\times 2}^*(4,2)\\
z^+\sin2\theta &&-z^--z^+\cos2\theta \\
z^--z^+\cos2\theta &&-z^+\sin2\theta \\
-2r\sin\theta && 2r\cos\theta\\
\color{red}A_{8\times 2}^*(8,1)&& \color{red}A_{8\times 2}^*(8,2)\\
\end{array}
\right]
\end{align}
where:
\begin{align*}
{\color{red}A_{8\times 2}^*(4,1)} &=2r{z_s}(\sigma\cos\theta+\tau\sin\theta)-z^-(\sigma{x_s}+\tau{y_s})-z^+((\tau{x_s}+\sigma{y_s})\sin2\theta+(\sigma{x_s}-\tau{y_s})\cos2\theta)\\
{\color{red}A_{8\times 2}^*(4,2)} &=2r{z_s}(\sigma\sin\theta-\tau\cos\theta)+z^-(\tau{x_s}-\sigma{y_s})+z^+((\tau{y_s}-\sigma{x_s})\sin2\theta+(\tau{x_s}+\sigma{y_s})\cos2\theta)\\
{\color{red}A_{8\times 2}^*(8,1)} &=z^+({y_s}\cos2\theta-{x_s}\sin2\theta)+2r{z_s}\sin\theta-z^-{y_s}\\
{\color{red}A_{8\times 2}^*(8,2)} &=z^+({x_s}\cos2\theta+{y_s}\sin2\theta)-2r{z_s}\cos\theta+z^-{x_s}
\end{align*}

Similarly, for the LC factorization of an affine camera in~\eqref{eqn:LC factorization affine}, if $\left(\pi_R\right)$ used is $\left(r\cos\theta,r\sin\theta,\sqrt{1-r^2}- 1,0\right)^T$, it will also lead to several differences. Here we only give the new $A_{6\times 2}$.
\begin{align}
A_{6\times 2}=\dfrac{1}{2} \left[
\begin{array}{ccc}
z^+\sigma+z^-(\sigma\cos{2\theta}+\rho\tau\sin{2\theta})&&z^-(\sigma\sin{2\theta}-\rho\tau\cos{2\theta})-z^+\rho\tau\\
z^+\rho\tau+z^-(\sigma\sin{2\theta}-\rho\tau\cos{2\theta})&& z^+\sigma-z^-(\sigma\cos{2\theta}+\rho\tau\sin{2\theta})\\
-2r(\sigma\cos{\theta}+\rho\tau\sin{\theta})&&2r(\rho\tau\cos{\theta}-\sigma\sin{\theta})\\
z^-\rho\sin{2\theta}&&-\rho\left(z^-\cos{2\theta}+z^+\right)\\
\rho\left(z^+-z^-\cos{2\theta}\right)&&-z^-\rho\sin{2\theta}\\
-2r\rho\sin{\theta}&&2r\rho\cos{\theta}\\
\end{array}
\right]
\end{align}

Note that when the image plane $(\pi)$ is exactly the $xoy$ axis plane $(0,0,1,0)^T$, when the above option on the mirror plane is chosen, for both pinhole~\eqref{LC camera representation0} and affine~\eqref{eqn:LC factorization affine} cameras, the {\em reflection} matrix should be an identity matrix instead of in the Housholder elementary matrix form(Table~\ref{classification table}) per Lemma~\ref{lemma:bisection planes}.
\section*{Acknowledgement}
{\small anonymous}
{\footnotesize
\bibliographystyle{ieee}
\bibliography{triangulation}
}

\def\arraystretch{1.65}

\begin{sidewaystable}
\centering
\caption{\large Classification and Definitions of Geometric Transformations Which are Stereohomology}
\label{classification table}
\resizebox{23cm}{!} {
 \begin{tabular}[t]{ccccccrr}%
\hlinewd{2pt}
\; No.\; & \tc{$S$ vs. $\pi$} & \tc{ \(\displaystyle\begin{casesBig} {\text{ Transformation}}{\text { matrix property}}\end{casesBig}\)} & \tc{Property of $\pi$} & \tc{ Property of $S$ }& {Transformation matrix formula\(\displaystyle(\lambda=1)\)} & \tc{Definition of transformation} \tabularnewline
\hline
1& $S\notin\pi$ & \tc{Singular} & \tc{Ordinary} & \tc{Ordinary} & & Central Projection \tabularnewline
2& $S\notin\pi$ & \tc{Singular} &\tc{Ordinary} & \tc{Infinite} &   \(\displaystyle\mathscr{T}\left ((s),(\pi); \lambda \right )=
 \lambda\cdot{\boldsymbol I} -\lambda\cdot \frac{{(s)\cdot(\pi) ^{\scriptscriptstyle \top}}}{{(s)^{\scriptscriptstyle \top}  \cdot (\pi)} } \)  & Oblique \& Orthographic Parallel Projection \tabularnewline
3& $S\notin\pi$ & \tc{Singular} & \tc{Infinite} & \tc{Ordinary} & & Direction  \tabularnewline
\hline
4& $S\notin\pi$ & \tc{Nonsingular} & \tc{Ordinary} & \tc{Ordinary} & & {Space homology}\tabularnewline
5& $S\notin\pi$ & \tc{Nonsingular} & \tc{Ordinary} & \tc{Infinite} &  \(\displaystyle\mathscr{T}(\ (s),(\pi) ;\rho, \lambda)\:\mathop =\;{\lambda\cdot\boldsymbol I} + (\rho - \lambda) \cdot \frac{{(s)\cdot(\pi) ^{\scriptscriptstyle \top} }} {{(s)^{\scriptscriptstyle \top} \cdot (\pi) }} \) & Oblique \& Orthographic Elementary Scaling  \tabularnewline
6& \tc{$S\notin\pi$} & \tc{Nonsingular} & \tc{Infinite} & \tc{Ordinary} & & Central Dilation  \tabularnewline
\hline
7& $S\notin\pi$ & \tc{$\begin{casesBig} {\textrm{ Nonsingular}}{\textrm {\&Involutory}}\end{casesBig}$} & \tc{Ordinary} & \tc{Ordinary} & & {Involutory space homology} \tabularnewline
8& $S\notin\pi$ & \tc{$\begin{casesBig} {\textrm {Nonsingular}}{\textrm {\&Involutory}}\end{casesBig}$} & \tc{Ordinary} & \tc{Infinite} & \(\displaystyle\mathscr{T}\left ( (s),(\pi); \lambda \right )=
 \lambda\cdot{\boldsymbol I} - 2\lambda \cdot {\huge \frac{(s)\cdot(\pi)\! ^{\scriptscriptstyle\top}} {(s)\!^{\scriptscriptstyle\top} \! \cdot (\pi) }} \) & Skew(Oblique) \& Orthographic Reflection \tabularnewline
9& $S\notin\pi$ & \tc{$\begin{casesBig} {\textrm {Nonsingular}}{\textrm {\&Involutory}}\end{casesBig}$} & \tc{Infinite} & \tc{Ordinary} & & Central Symmetry \tabularnewline
\hline
10& $S\in\pi$ & \tc{Nonsingular} & \tc{Ordinary} & \tc{Ordinary} & & {Space elation}\tabularnewline
11& $S\in\pi$ & \tc{Nonsingular} & \tc{Ordinary} & \tc{Infinite} &  \(\displaystyle \mathscr{T}((s),(\pi) ;\lambda,\mu )=
{\lambda\cdot\boldsymbol I} + \frac{ \mu \cdot
(s)\cdot{(\pi)}^{\scriptscriptstyle \top} } {\sqrt
{(s)^{\scriptscriptstyle \top}\!\!
\cdot\! (s)\!\cdot\!(\pi) ^{\scriptscriptstyle \top}\!\!\cdot\! (\pi) } }\) & Shearing  \tabularnewline
12& $S\in\pi$ & \tc{Nonsingular} & \tc{Infinite} & \tc{Infinite} & & Translation  \tabularnewline
\hlinewd{1pt}
\end{tabular}
}
\end{sidewaystable}

\end{document}